\PassOptionsToPackage{table}{xcolor}
\documentclass[10pt,twocolumn,letterpaper]{article}

\usepackage[pagenumbers]{cvpr} 

\usepackage[T1]{fontenc}
\usepackage{cuted}
\usepackage{microtype}
\usepackage{nicefrac}
\usepackage{multirow}
\usepackage{makecell}
\usepackage{array}
\usepackage{threeparttable}
\usepackage{wrapfig}
\usepackage{needspace}
\usepackage{tcolorbox}
\usepackage{float}
\usepackage{placeins}

\usepackage{amsmath,amsfonts,bm}

\def\eqref#1{equation~\ref{#1}}

\def\1{\bm{1}}

\DeclareMathAlphabet{\mathsfit}{\encodingdefault}{\sfdefault}{m}{sl}
\SetMathAlphabet{\mathsfit}{bold}{\encodingdefault}{\sfdefault}{bx}{n}

\newcommand{\method}{MomentMem\xspace}
\newcommand{\benchmark}{VMBench\xspace}

\makeatletter
\def\cvprsection{\@startsection {section}{1}{\z@}
   {-10pt plus -2pt minus -2pt}{7pt} {\large\bf\raggedright\hyphenpenalty\@M\exhyphenpenalty\@M}}
\def\cvprsubsection{\@startsection {subsection}{2}{\z@}
   {-8pt plus -2pt minus -2pt}{5pt} {\elvbf\raggedright\hyphenpenalty\@M\exhyphenpenalty\@M}}
\makeatother

\definecolor{cvprblue}{rgb}{0.21,0.49,0.74}
\usepackage[pagebackref,breaklinks,colorlinks,allcolors=cvprblue,hypertexnames=false]{hyperref}

\def\confName{CVPR}
\def\confYear{2027}

\title{\method: Memorizing Key Moments for Robot Manipulation}
\author{
Guangyu Zhao$^{1*}$ \quad
Yongchang Zhang$^{1*}$ \quad
Ruizhang Jiang$^{1*}$ \quad
Zhihong Yu$^{3}$ \quad
Yuhao Zhang$^{1}$ \\
Tao Wu$^{4}$ \quad
Qimin Zhou$^{5}$ \quad
Kunsong Shi$^{2}$ \quad
Haoran Wen$^{2}$ \\
Zeyu Zhang$^{6}$ \quad
Yin Li$^{7}$ \quad
Kun Zhan$^{2}$ \quad
Yiwu Zhong$^{1}$ \\[0.5em]
{\small
$^{1}$State Key Laboratory of General Artificial Intelligence, Peking University \quad
$^{2}$Li Auto Inc. \quad
$^{3}$Tsinghua University} \\
{\small
$^{4}$Nanyang Technological University \quad
$^{5}$Beijing University of Posts and Telecommunications} \\
{\small
$^{6}$Beijing Institute for General Artificial Intelligence \quad
$^{7}$University of Wisconsin - Madison} \\[0.3em]
{\small $^{*}$Equal contribution}
}

\begin{document}
\maketitle
\begin{strip}
\centering
\includegraphics[width=\textwidth]{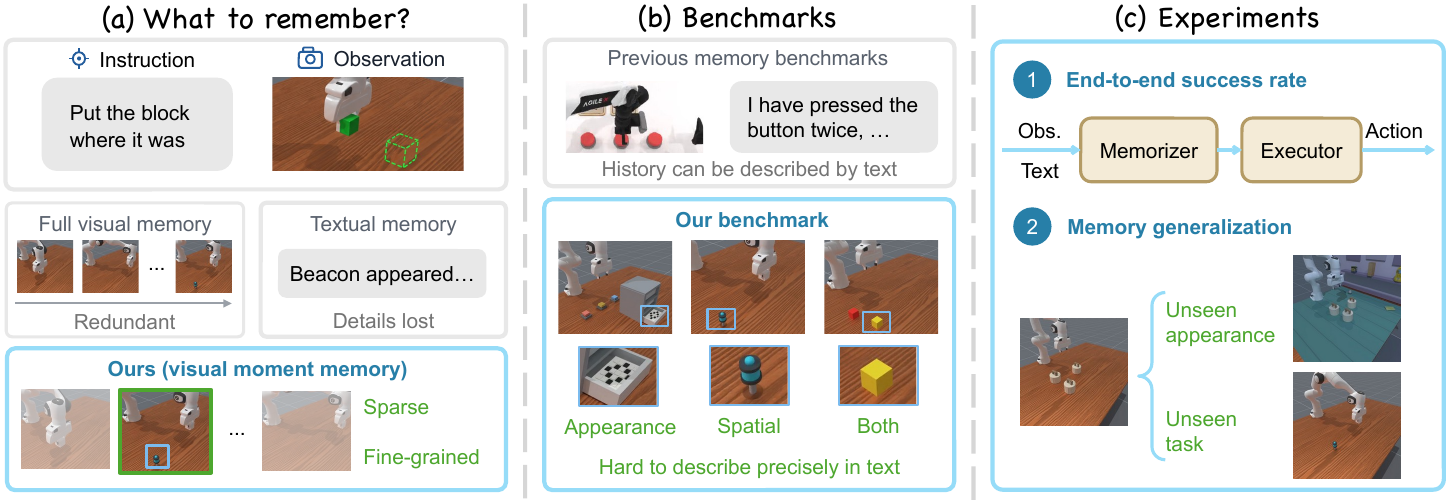}
\captionsetup{hypcap=false}
\captionof{figure}{\textbf{Overview of \benchmark and \method.} \textbf{(a)} Robots need historical information that is no longer visible, such as an object's previous location. Text summaries may omit precise visual details, while the full image history contains redundancy. We use sparse keyframes to retain raw visual evidence from key moments. \textbf{(b)} \benchmark focuses on visual moment memory for briefly visible appearances, spatial locations, and their associations. \textbf{(c)} \method decouples the memorizer from the executor. Experiments evaluate the complete policy's task success and study the memorizer's transfer to unseen appearances and few-shot adaptation to a held-out task.}
\label{fig:teaser}
\end{strip}
\begin{abstract}
Robotic manipulation often requires remembering visual information that is no longer observable, such as the original location of an object. 
Existing approaches typically rely on full observation histories, which become redundant over long horizons, or text summaries, which may omit precise visual details. 
We study \textbf{visual moment memory}, the ability to retain such details from key moments for subsequent manipulation. 
To support this study, we introduce \benchmark, a suite of long-horizon manipulation tasks requiring memory of spatial locations, visual appearances, and their associations. 
We further propose \method, a hierarchical architecture in which a VLM-based memorizer selects keyframes and an executor predicts actions conditioned on the selected memories. 
This separation enables us to evaluate memory selection alongside closed-loop task success and to study memorizer generalization independently of the executor. 
On \benchmark, \method achieves 49.9\% average success, outperforming the strongest evaluated baseline by \textbf{31.7 percentage points}. 
Moreover, we find that the memorizer \textbf{transfers well} to unseen appearances zero-shot. It can also adapt to an unseen task with only a few memory annotations, reaching 43.7\% success, \textbf{close to} the 44.3\% obtained with full memory supervision. 
Further experiments show competitive performance on RMBench and improvements in real-robot manipulation. These results highlight the value of retaining sparse visual evidence for memory-dependent manipulation.
Our code is available at \url{https://github.com/PKU-VaLuE-Lab/MomentMem}.
\looseness=-1\par
\end{abstract}

\section{Introduction}

Recently, memory has increasingly become a key capability of multi-modal models~\citep{moviechat,malmm,flashvstream,fluxmem} and embodied models~\citep{memoryvla,pam,memer,bpp}. Despite advances in embodied models~\citep{openvla,pi05,geminirobotics}, memory for robot manipulation remains challenging. As shown in Figure~\ref{fig:teaser}(a), existing approaches often focus on full observation histories~\citep{nativemem} or text descriptions~\citep{rmbench,pimem}.
Full observation histories are complete yet redundant, thus falling short when facing long-horizon tasks~\citep{eventvla,fluxmem}.
Text captures abstract events, yet can omit visual details, or cannot fully describe certain information (e.g., ``move the cube back to where it was'').
To this end, we study \textbf{visual moment memory} for robotic manipulation. It requires models to remember precise visual information at a key moment, such as texture patterns, spatial relations, and spatial locations. This type of memory is difficult to preserve in text.\looseness=-1

As a first step, we propose \benchmark, a benchmark of long-horizon manipulation tasks that require visual moment memory.
Its key design principle is to make successful manipulation depend on remembering precise visual information.
Different from existing benchmarks~\citep{mikasa,robomme,rmbench,robomemarena}, our benchmark emphasizes tasks that cannot be well addressed by concise text memory.
Specifically, these tasks require remembering exact positions and visual patterns, rather than only object identities or task progress.
Concise text summaries may omit these details, as illustrated in Figure~\ref{fig:teaser}(b).
We organize the tasks around spatial locations, visual appearances, and appearance-location bindings.
For example, returning a cube to its original position requires remembering exactly where it was, rather than just knowing that it was moved.
Across these tasks, the key visual moment is only visible across a short period of time and the model has to capture such moments for successful task completion. 
Together, these tasks provide a testbed for studying how robots memorize precise visual information for manipulation tasks.\looseness=-1

Going further, we consider the modeling of visual moment memory for robot manipulation.
Task success alone does not reveal whether a failure comes from memory or execution~\citep{memobench}.
Beyond overall task performance, we seek to examine two key capabilities: memory selection and action prediction.
To this end, we propose \method, a hierarchical architecture that separates memory from action prediction.
This separation allows us to examine memory selection and action execution independently.
As shown in Figure~\ref{fig:teaser}(c) (top), a VLM memorizer selects keyframes, and then the executor predicts actions.
We observe that the key event often spans across several frames that show the same visual cue.
We annotate this span as a temporal segment.
We aim for the memorizer to select a keyframe in the segment as early as possible.
Also, if it misses an earlier frame, it should still try to capture the later frame in the same segment.
During inference, the executor predicts actions from the selected keyframes together with the current observation.
Details are provided in Section~\ref{sec:methodology}.\looseness=-1

With our benchmark tasks and hierarchical model, we design metrics and protocols to evaluate end-to-end success rate and the generalization of memory capability, as shown in Figure~\ref{fig:teaser}(c).
As in most existing benchmarks, we test the models on our benchmark tasks and measure the success rate over closed-loop rollouts.
We evaluate how well the memorizer captures the relevant evidence.
We also test its transfer to unseen appearances and adaptation to new tasks.
As shown in Figure~\ref{fig:teaser}(c) (bottom), we vary the visual appearance and directly test the overall model, with the memorizer trained only on the original visual appearances.
Besides, we test the overall model for unseen tasks, allowing the memorizer to adapt to new tasks with a few memory annotations.
In both protocols, we train the executor on target-domain action demonstrations using memories predicted by the memorizer.
In other words, our protocols measure the generalization of the memorizer rather than the executor.\looseness=-1

On \benchmark, \method achieves \textbf{49.9\% average success}, exceeding the strongest evaluated baseline by \textbf{31.7 percentage points}.
In addition, the memorizer transfers \textbf{zero-shot} to unseen appearances.
With only \textbf{a few memory annotations}, it adapts to an unseen task.
The resulting policy reaches \textbf{43.7\% success}, close to 44.3\% with full memory supervision.
Beyond \benchmark, \method remains competitive on RMBench and improves task success in real-robot manipulation.
These results demonstrate the effectiveness and generalization of \method.\looseness=-1

Our contributions are as follows:
\begin{itemize}[leftmargin=2.0em, itemsep=2pt, topsep=2pt]
    \item We identify visual moment memory as a key research problem for robotic manipulation, focusing on remembering transient, precise visual information.
    \item We introduce \benchmark to evaluate and study this problem in closed-loop manipulation tasks. We also propose \method, a hierarchical model that supports separate diagnosis of memory and action prediction.
   \item Extensive experiments show that \method exceeds the strongest baseline on our \benchmark by 31.7 percentage points. Our analysis demonstrates that the memorizer can transfer to unseen appearance and task. Further, \method performs well on RMBench and real-robot tasks.\looseness=-1
\end{itemize}

\section{Related Work}
\subsection{Vision-Language-Action Models}

Vision-Language-Action (VLA) models leverage pretrained vision-language representations to generate robot actions. RT-2~\citep{rt2} transferred web-scale semantic knowledge to robotic control, while OpenVLA~\citep{openvla} and Octo~\citep{octo} developed open generalist policies trained on diverse robot data. Recent models, including $\pi_0$~\citep{pi0}, $\pi_{0.5}$~\citep{pi05}, GR00T series~\citep{grootn1}, GigaBrain series~\citep{gigabrain0,gigabrain05,gigabrain07}, and Gemini Robotics~\citep{geminirobotics}, improve action generation and cross-task generalization.\looseness=-1

Despite these advances, most VLAs condition on the current observation or a short context window. They therefore struggle when relevant evidence has disappeared from view~\citep{pi0,memoryvla,kcvla,muvla}. Our work addresses this limitation by providing VLA models with task-relevant keyframes identified from earlier observations.\looseness=-1

\subsection{Robotic Manipulation Policies with Memory}

Some robotic manipulation tasks require evidence from earlier observations, motivating policies that use interaction history~\citep{sam2act,memoryvla,rmbench}.
Available information includes the task instruction, visual observations, actions, and proprioceptive states.
Existing memory methods differ in how they store and use historical information.
Some methods maintain recurrent states, memory banks, or compressed tokens to retain information from long histories~\citep{memoryvla,memoryvlapp,muvla,rememvla,rbvla,nativemem,echo,pam}.
Compressing long histories may discard visual details needed for later actions.
Other methods summarize the history with language or use textual plans as memory~\citep{rmbench,memer,pimem}.
Such representations are compact, but concise text summaries may omit task-relevant visual details, such as fine-grained appearances.
Another line of work directly selects observations from the visual history~\citep{bpp}.
This preserves raw visual evidence, though several methods select with predefined rules or heuristics~\citep{kcvla,helm,kemo,robomemarena}.
Such rules may need adjustment when memory requirements change.\looseness=-1

Our method learns when to save observations that contain evidence needed for later actions.
It uses a hierarchical architecture with separate memorizer and executor modules.
The most closely related method is EventVLA~\citep{eventvla}, which predicts keyframes from predicted future actions within a single model.
Our memorizer instead selects keyframes from visual observations independently of action prediction.\looseness=-1

\subsection{Robotic Memory Benchmarks}

Several benchmarks evaluate memory in robotic manipulation.
RMBench~\citep{rmbench} supports systematic evaluation of memory in robotic imitation learning.
MIKASA-Robo~\citep{mikasa} and RoboMME~\citep{robomme} evaluate multiple dimensions of memory, while MemoryBench~\citep{sam2act} focuses on spatial memory and action recall.
Other benchmarks focus on memory in long-horizon tasks, which involve extended trajectories and require precise manipulation~\citep{robomemarena,eventvla,embodiedslotssm,prism}.
Complementing these benchmarks, \benchmark focuses on visual moment memory: recalling precise visual evidence that is visible only for a short period of time.
We study memorizer transfer using visual OOD variants and few-shot adaptation on a held-out task.\looseness=-1

\section{\texorpdfstring{\benchmark: Evaluating Visual Moment Memory}{VMBench: Evaluating Visual Moment Memory}}
\label{sec:benchmark}

We introduce \benchmark that necessitates visual moment memory during robot manipulation.
It contains nine closed-loop manipulation tasks, requiring the model to recall precise visual details which text summaries are hard to describe.
This evidence is briefly presented by the environment or revealed through interaction.
In some tasks, the environment presents the evidence before execution.
In others, the robot must act to reveal it, such as opening drawers or scanning with a detector.
These settings test memory during both observation and active manipulation.
Appendix~\ref{app:benchmark} describes the key evidence in each task, and appendix~\ref{app:task-instructions} provides complete task instructions for all tasks in \benchmark.

\begin{table*}[!t]
\caption{\textbf{Overview of the tasks in \benchmark.} Tasks are grouped by the visual information they require: spatial locations, appearances, or their associations. Image sequences illustrate how evidence is acquired and used after it is no longer visible. The final column shows each task's visual OOD variant.}
\label{tab:benchmark-overview}
\centering
\begingroup
\fontsize{8}{9.3}\selectfont
\setlength{\tabcolsep}{6pt}
\renewcommand{\arraystretch}{1.05}
\setlength{\extrarowheight}{0pt}
\newlength{\BTimagewidth}
\newlength{\BTimagegap}
\newlength{\BTsequencewidth}
\newlength{\BTcategorywidth}
\newlength{\BTtaskwidth}
\newlength{\BTdescriptionwidth}
\setlength{\BTimagewidth}{3\baselineskip}
\setlength{\BTimagegap}{2pt}
\setlength{\BTsequencewidth}{\dimexpr4\BTimagewidth+3\BTimagegap\relax}
\setlength{\BTcategorywidth}{52pt}
\setlength{\BTtaskwidth}{58pt}
\setlength{\BTdescriptionwidth}{\dimexpr\textwidth-8\tabcolsep-\BTcategorywidth-\BTtaskwidth-\BTsequencewidth-\BTimagewidth\relax}
\newcommand{\BTtask}[2]{\textbf{#1}\newline\textbf{#2}}

\begin{tabular}{@{}
  >{\centering\arraybackslash}m{\BTcategorywidth}
  >{\raggedright\arraybackslash}m{\BTtaskwidth}
  >{\raggedright\arraybackslash}m{\BTdescriptionwidth}
  >{\raggedright\arraybackslash}m{\BTsequencewidth}
  >{\centering\arraybackslash}m{\BTimagewidth}
@{}}
\toprule
\textbf{Category} & \textbf{Task} & \textbf{Description} &
\multicolumn{1}{c}{\textbf{Task sequence}} & \textbf{Visual}\newline\textbf{OOD} \\
\midrule
& \BTtask{Spring}{Compression}
& Compress the spring back to its deepest previously reached position.
& \makebox[\linewidth][l]{%
    \includegraphics[width=\BTimagewidth]{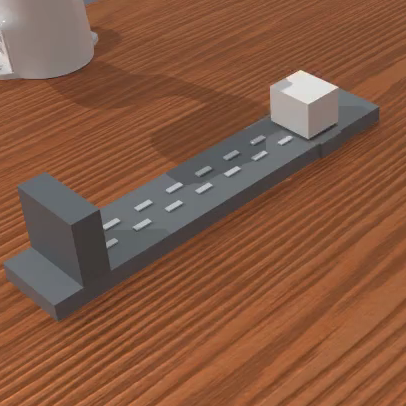}\hspace{\BTimagegap}%
    \includegraphics[width=\BTimagewidth]{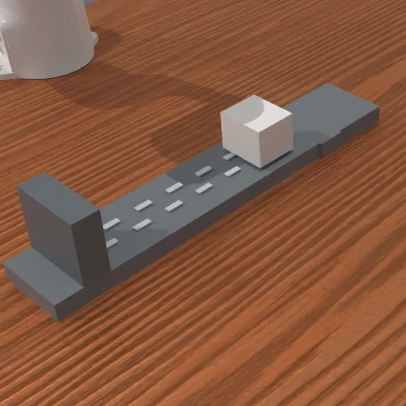}\hspace{\BTimagegap}%
    \includegraphics[width=\BTimagewidth]{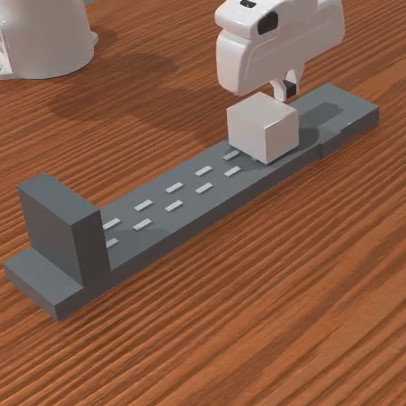}%
  }
& \includegraphics[width=\BTimagewidth]{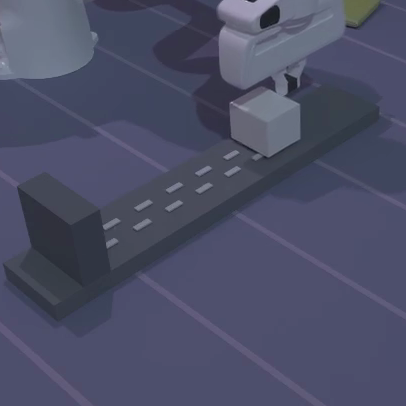} \\[3pt]
\textbf{Spatial}
& \BTtask{Flash}{Beacon}
& Move the cube to the location where the beacon briefly appeared.
& \makebox[\linewidth][l]{%
    \includegraphics[width=\BTimagewidth]{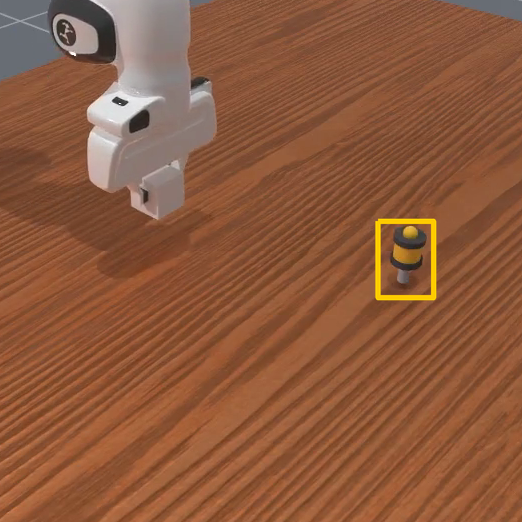}\hspace{\BTimagegap}%
    \includegraphics[width=\BTimagewidth]{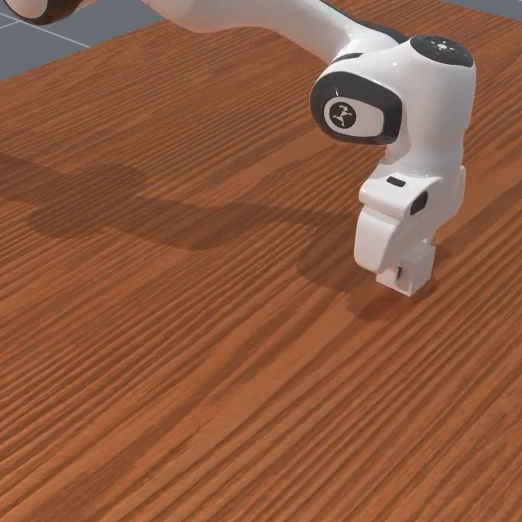}\hspace{\BTimagegap}%
    \includegraphics[width=\BTimagewidth]{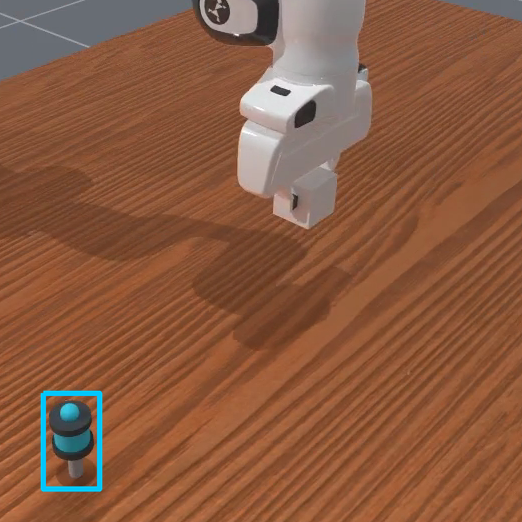}\hspace{\BTimagegap}%
    \includegraphics[width=\BTimagewidth]{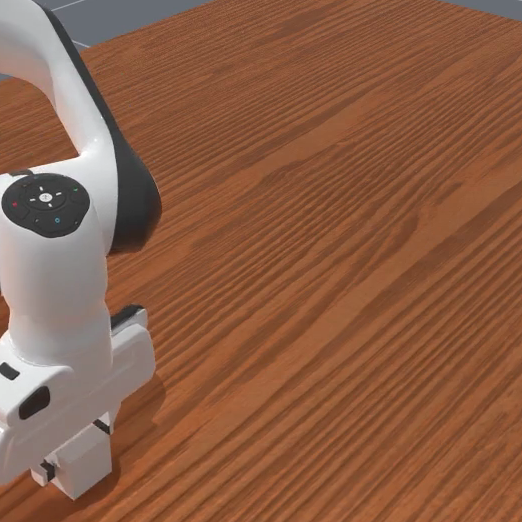}%
  }
& \includegraphics[width=\BTimagewidth]{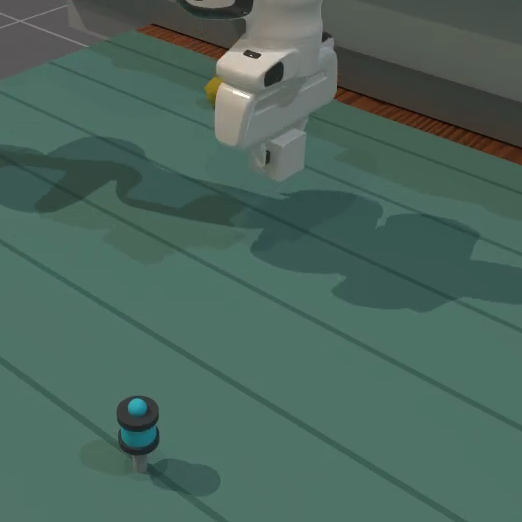} \\[3pt]
& \BTtask{Explore}{Cube}
& Search all the cups and retrieve the hidden cube.
& \makebox[\linewidth][l]{%
    \includegraphics[width=\BTimagewidth]{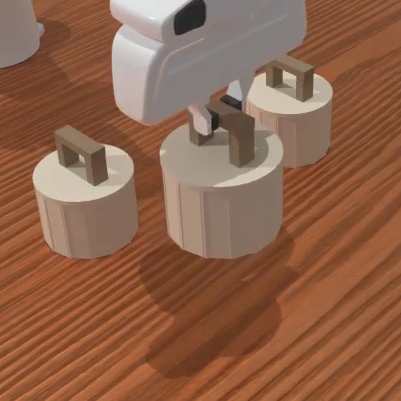}\hspace{\BTimagegap}%
    \includegraphics[width=\BTimagewidth]{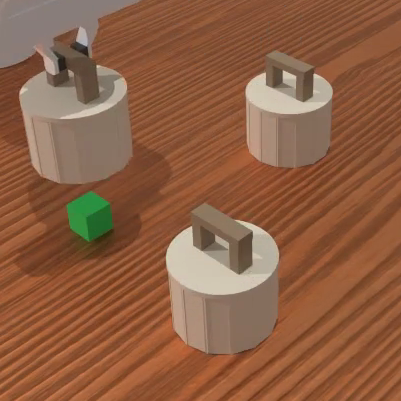}\hspace{\BTimagegap}%
    \includegraphics[width=\BTimagewidth]{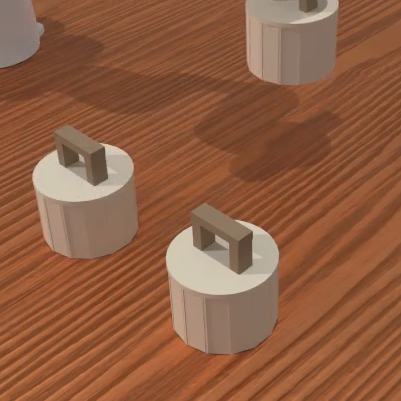}\hspace{\BTimagegap}%
    \includegraphics[width=\BTimagewidth]{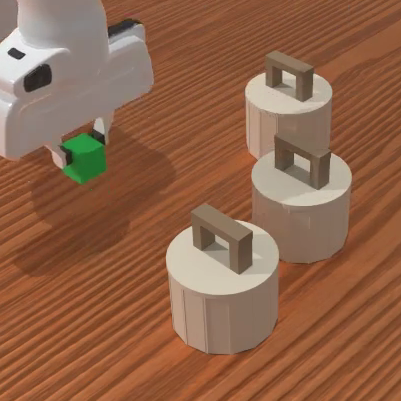}%
  }
& \includegraphics[width=\BTimagewidth]{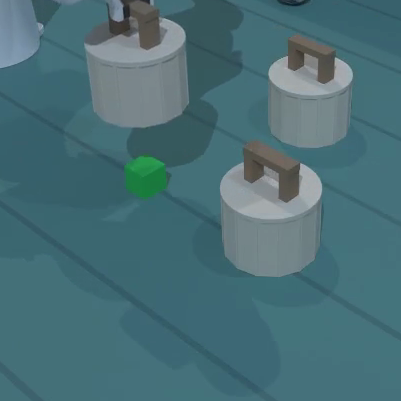} \\
\midrule
& \BTtask{Hidden}{Pattern}
& Push the box whose hidden pattern matches the pattern on the card.
& \makebox[\linewidth][l]{%
    \includegraphics[width=\BTimagewidth]{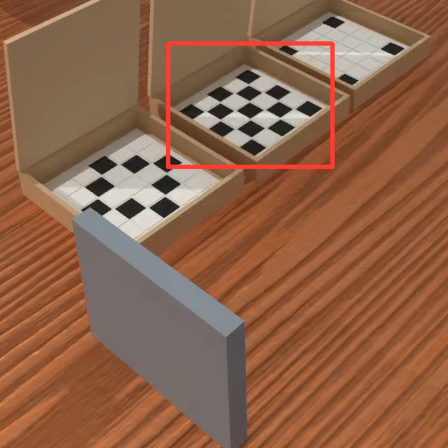}\hspace{\BTimagegap}%
    \includegraphics[width=\BTimagewidth]{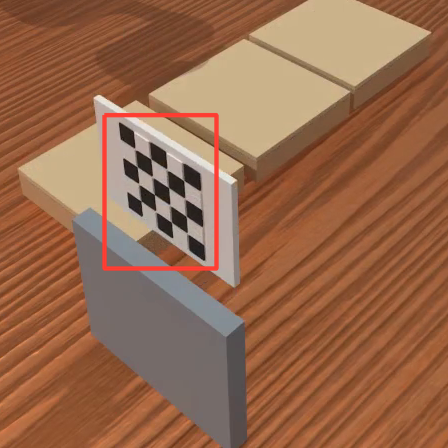}\hspace{\BTimagegap}%
    \includegraphics[width=\BTimagewidth]{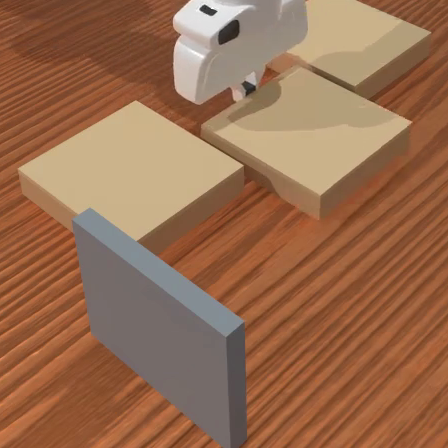}%
  }
& \includegraphics[width=\BTimagewidth]{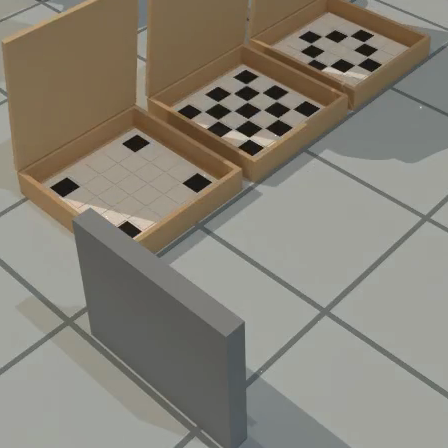} \\[3pt]
\textbf{Appearance}
& \BTtask{Drawer}{Match}
& Open the upper or middle drawer whose pattern matches that of the bottom drawer.
& \makebox[\linewidth][l]{%
    \includegraphics[width=\BTimagewidth]{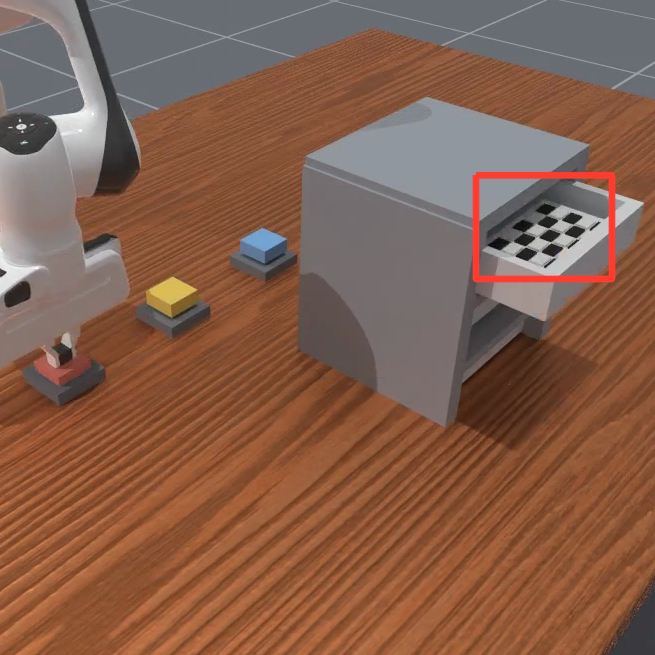}\hspace{\BTimagegap}%
    \includegraphics[width=\BTimagewidth]{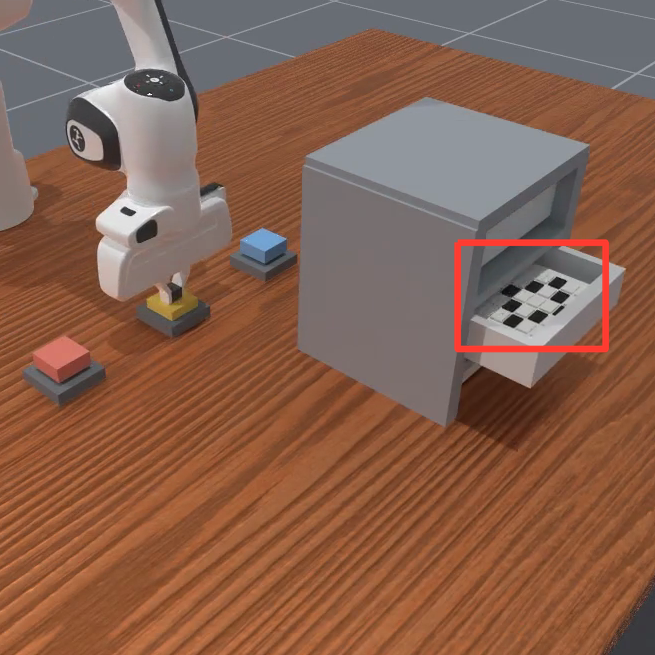}\hspace{\BTimagegap}%
    \includegraphics[width=\BTimagewidth]{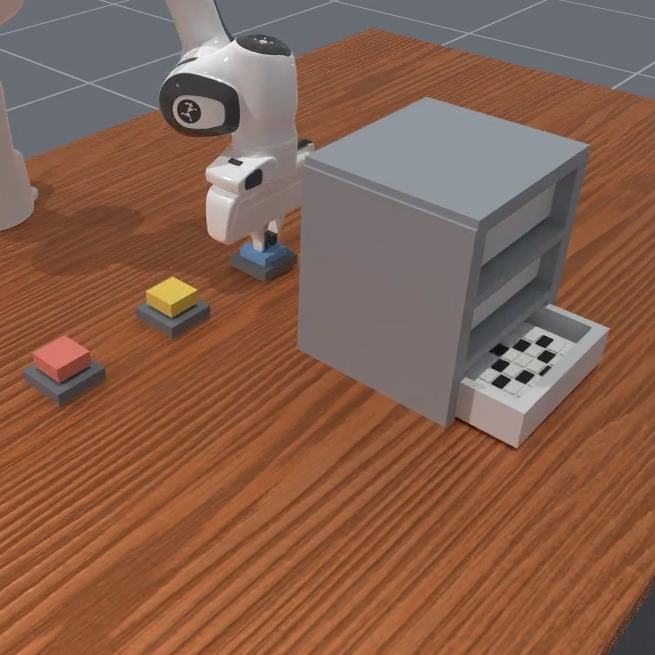}\hspace{\BTimagegap}%
    \includegraphics[width=\BTimagewidth]{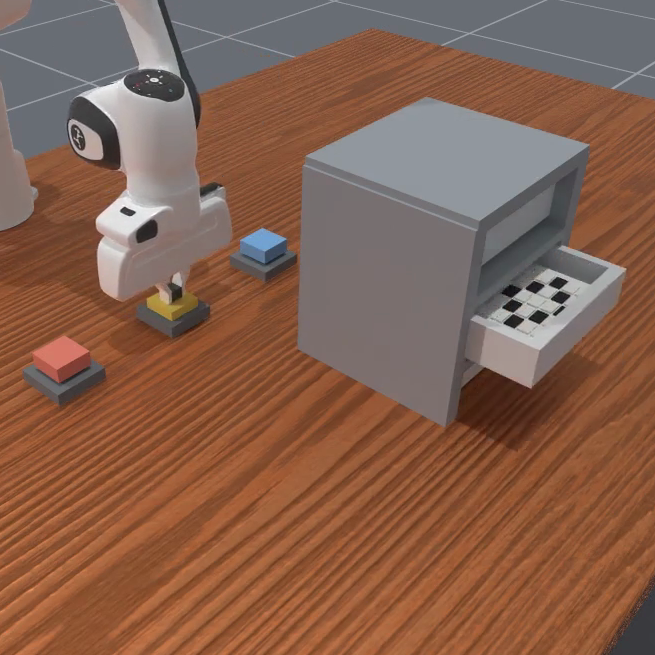}%
  }
& \includegraphics[width=\BTimagewidth]{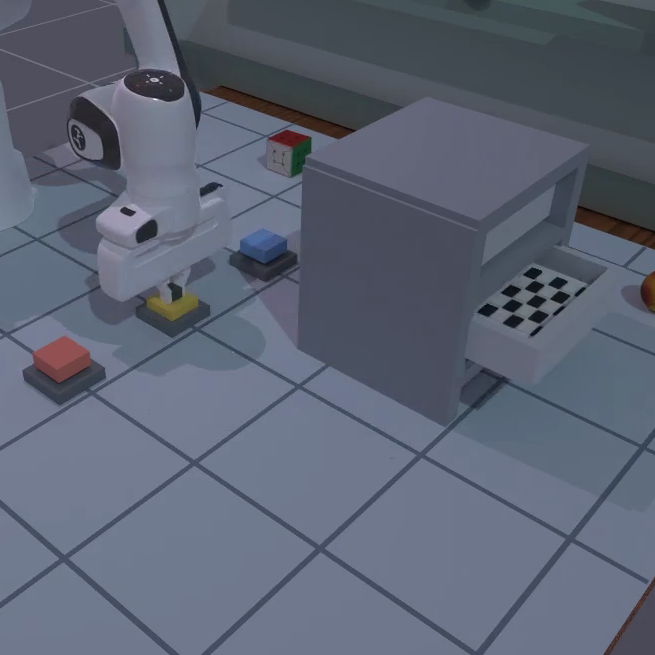} \\[3pt]
& \BTtask{Scroll}{Discovery}
& Reveal the scroll patterns, then stack the two matching scrolls.
& \makebox[\linewidth][l]{%
    \includegraphics[width=\BTimagewidth]{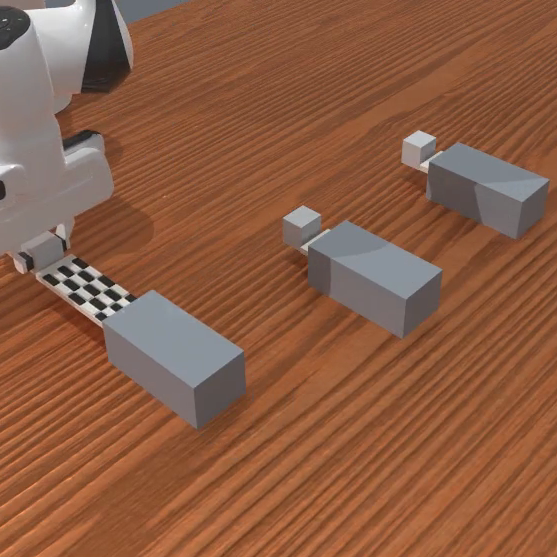}\hspace{\BTimagegap}%
    \includegraphics[width=\BTimagewidth]{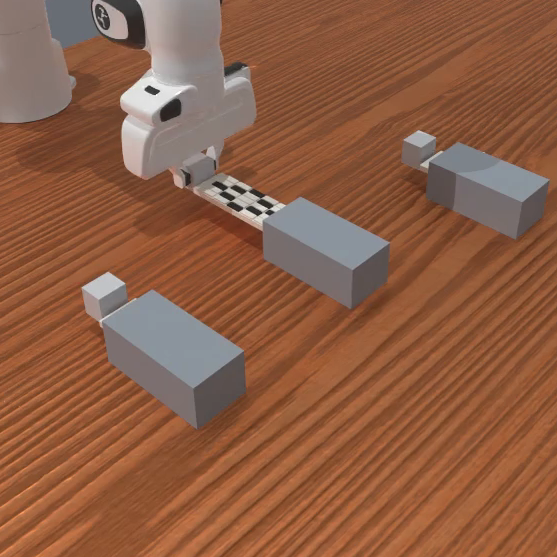}\hspace{\BTimagegap}%
    \includegraphics[width=\BTimagewidth]{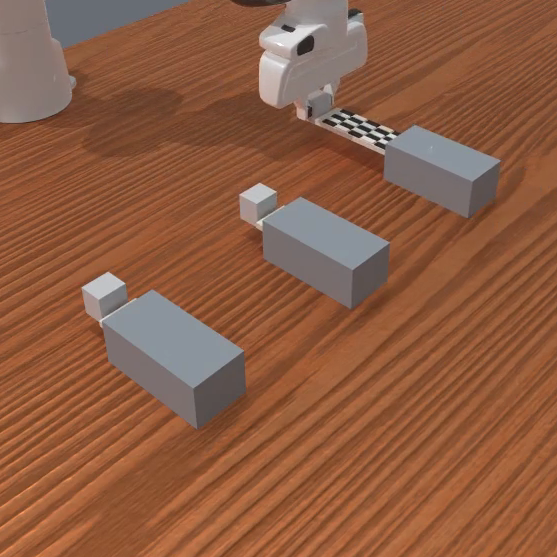}\hspace{\BTimagegap}%
    \includegraphics[width=\BTimagewidth]{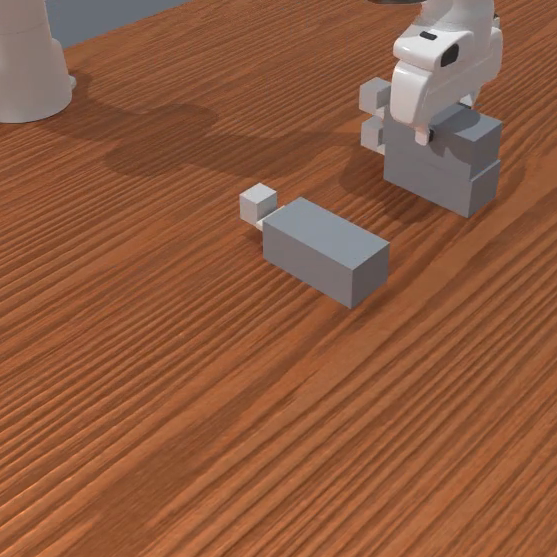}%
  }
& \includegraphics[width=\BTimagewidth]{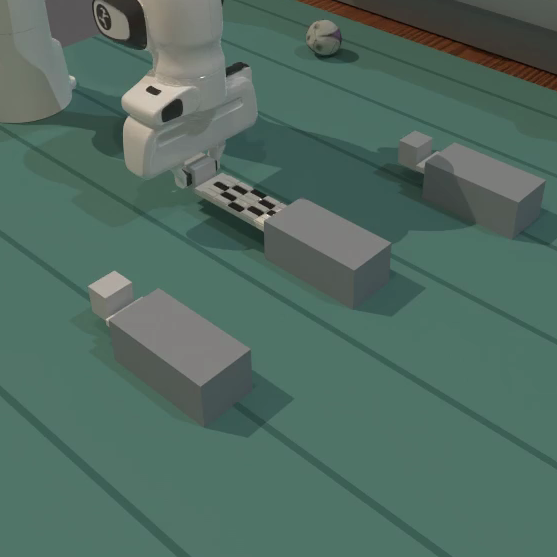} \\
\midrule
& \BTtask{Placement}{Imitation}
& Move each cube to its position shown in the demonstration.
& \makebox[\linewidth][l]{%
    \includegraphics[width=\BTimagewidth]{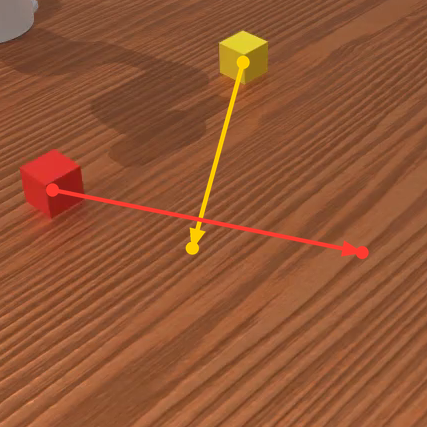}\hspace{\BTimagegap}%
    \includegraphics[width=\BTimagewidth]{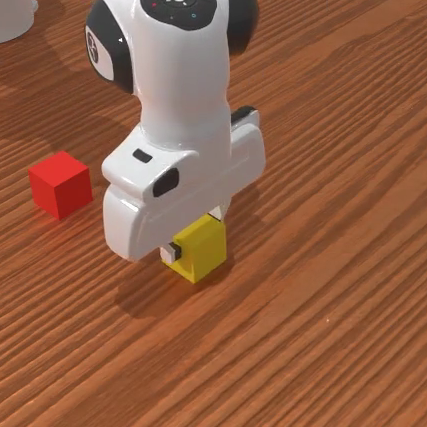}\hspace{\BTimagegap}%
    \includegraphics[width=\BTimagewidth]{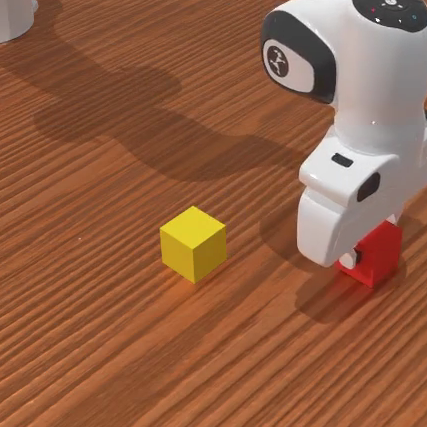}%
  }
& \includegraphics[width=\BTimagewidth]{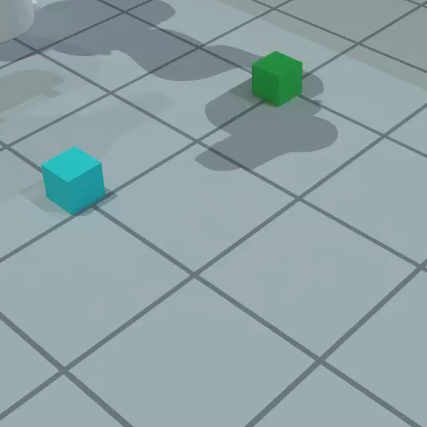} \\[3pt]
\textbf{Spatial \&}\newline\textbf{Appearance}
& \BTtask{Mine}{Sweeper}
& Place each cube where the detector briefly showed the same color.
& \makebox[\linewidth][l]{%
    \includegraphics[width=\BTimagewidth]{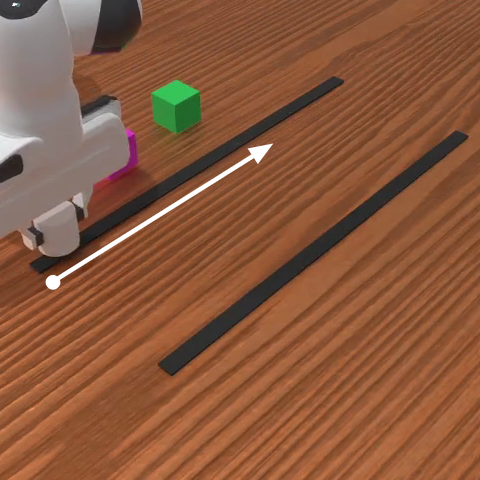}\hspace{\BTimagegap}%
    \includegraphics[width=\BTimagewidth]{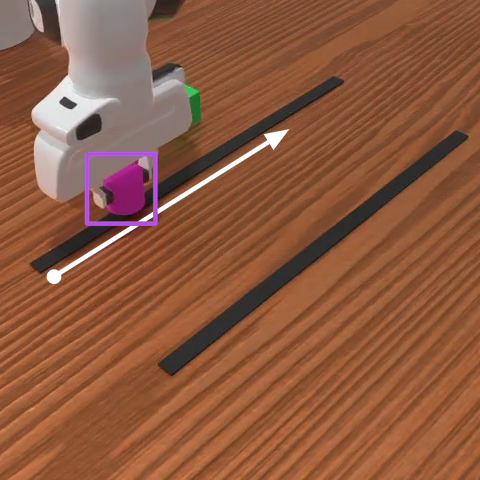}\hspace{\BTimagegap}%
    \includegraphics[width=\BTimagewidth]{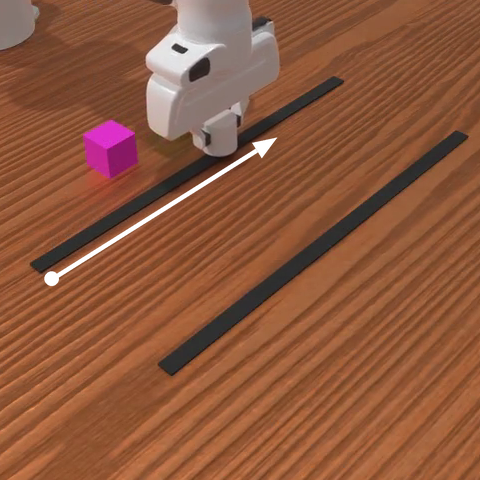}\hspace{\BTimagegap}%
    \includegraphics[width=\BTimagewidth]{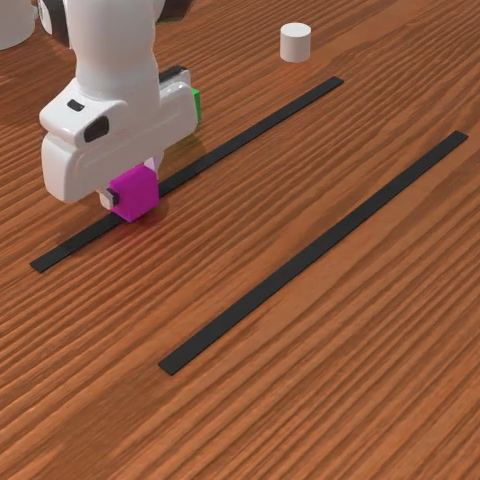}%
  }
& \includegraphics[width=\BTimagewidth]{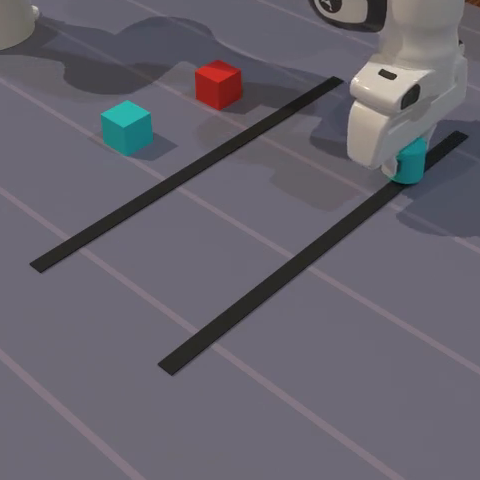} \\[3pt]
& \BTtask{Seal}{Checking}
& Lift the stamps, then place each cube on the matching revealed color.
& \makebox[\linewidth][l]{%
    \includegraphics[width=\BTimagewidth]{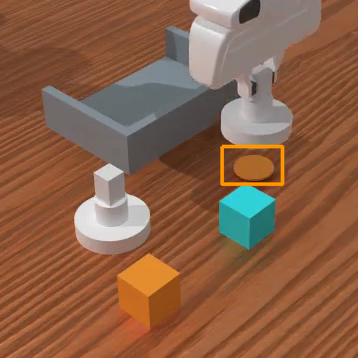}\hspace{\BTimagegap}%
    \includegraphics[width=\BTimagewidth]{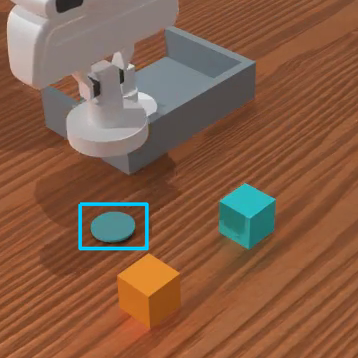}\hspace{\BTimagegap}%
    \includegraphics[width=\BTimagewidth]{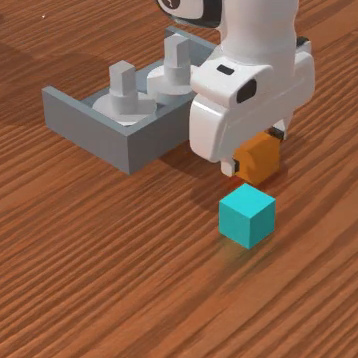}\hspace{\BTimagegap}%
    \includegraphics[width=\BTimagewidth]{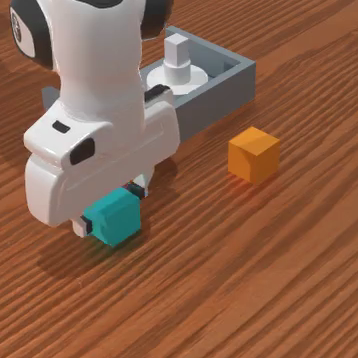}%
  }
& \includegraphics[width=\BTimagewidth]{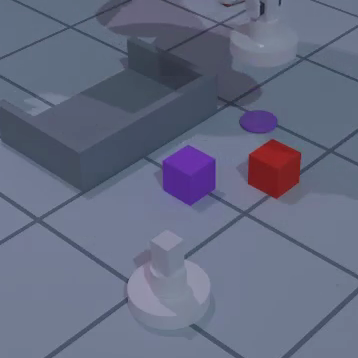} \\
\bottomrule
\end{tabular}
\endgroup
\end{table*}

\subsection{Types of Visual Evidence}

We categorize the tasks by their visual information requirements. Table~\ref{tab:benchmark-overview} provides an overview.

\textbf{Spatial locations.}
These tasks require recalling precise spatial locations.
The policy must retain a spatial reference for later movement or search.
For example, Flash Beacon requires the robot to place the cube where the beacon appeared.\looseness=-1

\textbf{Visual appearance.}
These tasks require retaining detailed visual patterns.
The policy must compare appearances observed at different times, even when object categories alone cannot distinguish them.
For example, Scroll Discovery requires the robot to retract scrolls, remembering the revealed patterns, and then stack the two scrolls with matching patterns after they roll back up.\looseness=-1

\textbf{Appearance-location bindings.}
These tasks require retaining associations between appearances and locations.
Remembering the appearances or locations separately is insufficient; the policy must preserve their correspondence.
For example, in Mine Sweeper, the robot scans the ``minefield'' with a detector.
The detector briefly changes color at each target location. After scanning, the robot places colored blocks at these locations, so it must remember where each color appeared.\looseness=-1

\subsection{Demonstrations and Memory Annotations}

We collect 100 successful demonstrations per task.
Each demonstration contains annotated key segments.
We obtain these segments from simulator events and task state.
Each segment marks when one piece of task-relevant evidence is visible.
It allows multiple valid frames rather than prescribing a single representative.
The annotations support both training and evaluation of keyframe selection.\looseness=-1

\benchmark is built on SAPIEN~\citep{sapien} and ManiSkill3~\citep{maniskill,maniskill2,maniskill3}. We collect demonstrations using a Franka Panda manipulator and scripted closed-loop experts. Trajectories record RGB images from overview and wrist cameras at $224\times224$ resolution, with a control frequency of 20 Hz. Privileged state is not included in policy observations. Appendix~\ref{app:data-construction} details data collection, randomization, and annotation.\looseness=-1

\subsection{Visual OOD Setting}

To test memory transfer across visual appearances, we provide a visual out-of-distribution (OOD) variant of each task.
We change room and table appearance, lighting, and distractors outside the manipulation area.
Task mechanics, memory events, and success conditions remain unchanged.\looseness=-1

This setting tests whether a memorizer trained on standard appearances can directly generate useful memories under unseen appearances. The executor is trained on target-domain action demonstrations.

\subsection{Memory Evaluation}
We evaluate complete policies by closed-loop task success.
For methods that explicitly select keyframes, the interval annotations also support a separate evaluation of memory selection.
We replay recorded demonstrations, using each model's previous selections as its memory context.
Segment hit rate measures the fraction of intervals with at least one selected frame.
Segment accuracy requires exactly one selection per interval.
Non-keyframe accuracy measures the fraction of demonstrations with no selections outside the annotated intervals.
These metrics distinguish missed evidence, duplicate selections, and selections outside the intervals.
Appendix~\ref{app:keyframe-quality} gives the details and results.\looseness=-1\par

\section{\method}
\label{sec:methodology}

\subsection{Overview}

\begin{figure*}[!t]
\centering
\includegraphics[width=0.95\textwidth]{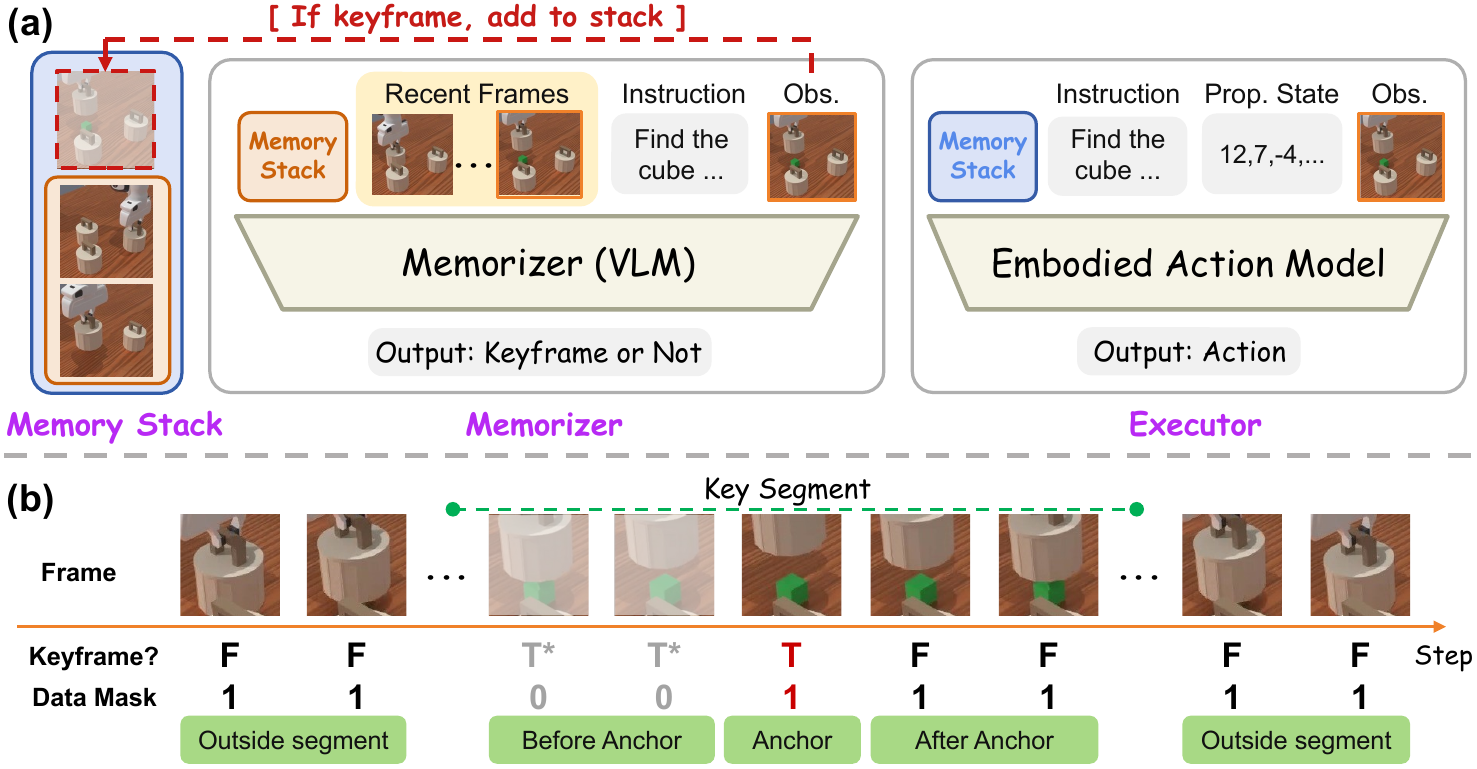}
\caption{\textbf{Overview of \method.} \textbf{(a)} The memorizer uses recent observations and existing memory to decide whether to store the current frame. The executor predicts actions from the updated memory and current input. \textbf{(b)} Each epoch, we sample an anchor from each key segment. Under a teacher-forced history, the anchor is positive. Earlier frames in the segment remain in the observation context but are excluded from the loss. Later frames are negative because the event has been stored.}
\label{fig:method}
\end{figure*}

As shown in Figure~\ref{fig:method}, \method uses a hierarchical architecture that decouples memory from action execution. A VLM memorizer processes streaming visual observations and selects keyframes. A VLA executor uses these keyframes and the current observation to predict actions.\looseness=-1

This design addresses two challenges of visual moment memory.
We use sparse keyframes as the interface between memory and execution.
Keyframes preserve raw images to retain visual details needed for later actions.
The memorizer uses recent observations and existing memory to decide when to save a frame. It captures transient evidence while avoiding duplicate storage.\looseness=-1

\textbf{Online interaction.} At time $t$, the robot receives an observation $o_t$, a proprioceptive state $s_t$, and a task instruction $l$. The memorizer $\pi_M$ decides whether to store the current frame using the existing memory $M_{t-1}$, a buffer $B_t$ of recent subsampled observations including $o_t$, and the instruction. The executor $\pi_E$ predicts an action chunk $A_t$ from the updated memory $M_t$ and the current input:
\begin{equation}
    \begin{aligned}
        p_t &= \pi_M(\text{Yes} \mid M_{t-1}, B_t, l), \\
        M_t &= M_{t-1} \cup \{o_t : p_t > 0.5\}, \\
        A_t &\sim \pi_E(\cdot \mid M_t, o_t, s_t, l).
    \end{aligned}
    \label{eq:memory-policy}
\end{equation}
The executor receives the selected keyframes in temporal order rather than the full observation history.

\textbf{Training pipeline.} We first train the memorizer with key segment annotations. We then freeze it to generate memories for action demonstrations. The executor takes the memorized keyframes as input and learns to predict the demonstration actions. The details are elaborated in the following subsections.

\subsection{The Memorizer of \method}
\label{sec:identifier-training}

We fine-tune a VLM to output ``Yes'' or ``No'' given the context in Equation~\ref{eq:memory-policy}, indicating whether to store the current frame. Training must allow multiple valid frames for each event and account for how earlier selections affect the current decision.\looseness=-1

\textbf{Segment-level supervision.}
\label{sec:keyframe-segments}
Visual evidence often remains visible across several frames.
Labeling only one fixed frame as correct would penalize other valid selections.
Our supervision instead allows different representatives of the same evidence across training epochs.
For each piece of visual evidence, we annotate an interval $\mathcal S_j=[u_j,v_j]$ during which it is visible.
Each epoch, we select one representative frame, called an anchor, from each interval.
The memorizer is trained to identify the anchors and reject others.
We generate intervals automatically in simulation, and annotate them manually for real-world experiments.\looseness=-1

\textbf{Anchor sampling.} At the start of each epoch, we sample one anchor $c_j$ per interval. Under a teacher-forced history, the anchor receives a positive label and enters memory. Earlier frames in the same interval are excluded from the loss but remain in the recent observation context. Later frames and frames outside all intervals receive negative labels.
Resampling across epochs teaches the model to select different valid representatives of the same event.
The memorizer is encouraged to capture visual evidence when it appears. If it misses an opportunity, it can select a later valid frame.
We use focal loss to address the imbalance between positive and negative samples.\looseness=-1

\textbf{Scheduled sampling.}
Whether a frame should be stored depends on what is already in memory.
The same evidence may need to be saved after a missed selection, but becomes redundant once a representative is stored.
Missed or extra selections during execution therefore change later inputs and decisions.
Training only with teacher-forced histories does not expose the memorizer to these cases.
We use sequence-level scheduled sampling to train with histories that include model predictions.
At the start of each epoch, the current model replays the training trajectories. We then mix teacher decisions and model predictions to construct memory histories. The probability of using model predictions increases during training. Supervision follows the history provided: after a missed anchor, later valid frames in the same interval remain positive. Once a representative is stored, subsequent frames become negative. Appendix~\ref{app:implementation} details label construction.\looseness=-1

\subsection{The Executor of \method}
\label{sec:executor-training}

The executor learns to act on the memories that the memorizer actually provides.
We therefore freeze the trained memorizer and replay action demonstrations in temporal order.
Its selections form the memory context for executor training.
The memorizer may select frames at different times, miss relevant evidence, or store extra frames.
Training with these predicted memories exposes the executor to the memorizer's outputs rather than only to frames sampled from annotated intervals.\looseness=-1

For a training sample at time $t$, the executor receives predicted keyframes from before $t$, the current observation $o_t$, the state $s_t$, and the instruction $l$. We supervise it with the demonstration action chunk using the original learning objective of the action policy.\looseness=-1

This interface supports separate adaptation of memory and execution. In the visual OOD experiments, we reuse the memorizer directly. For a new task, we first adapt it with a few memory annotations. In both settings, the executor is trained on target-domain action demonstrations. So, this hierarchical design allows us to evaluate the two modules~separately.\looseness=-1
\section{Experiments and Results}
\label{sec:experiments}

We first evaluate overall task success on \benchmark. We then test the memorizer's transfer to unseen appearances and adaptation to a new task. We find that \method improves closed-loop success on tasks that require visual moment memory. The memorizer also generalizes to unseen appearance and task to some extent. Beyond \benchmark, we test \method on RMBench and real-robot tasks. The results suggest that \method is broadly applicable to memory-dependent manipulation.

\subsection{Experimental Setup}

\textbf{Implementation and Evaluation.} 
We initialize the memorizer from Qwen3-VL-4B and the executor from $\pi_{0.5}$.
We use 100 demonstrations per task provided by \benchmark and jointly train all nine tasks.
We first train the memorizer with annotated key segments and then freeze it.
We use its predicted keyframes to train the executor on action demonstrations.
At test time, the executor receives online-selected keyframes, the current observation, the robot state, and the task instruction.
Appendix~\ref{app:implementation} provides the detailed training configuration.\looseness=-1

\textbf{Baselines.}
We compare \method against $\pi_{0.5}$~\citep{pi05} and three memory-augmented policies.
Mem-0~\citep{rmbench} uses explicit memory and high-level planning to guide action execution.
NativeMEM~\citep{nativemem} compresses past observations into memory tokens.
EventVLA~\citep{eventvla} learns sparse visual evidence memory jointly with action prediction.
All methods use the same demonstrations and joint training as \method.
Each task is evaluated on 100 rollouts with unseen initial configurations.\looseness=-1


\subsection{\method Improves Success on Visual Moment Memory Tasks}

\begin{table*}[!t]
\centering
\caption{\textbf{Performance comparison on \benchmark.} We compare \method with several baselines on \benchmark. All methods are trained on the same dataset, with 100 trajectories per task. We evaluate all methods under 100 unseen initialization conditions per task. For each method, the models are jointly trained and evaluated across all tasks.}
\label{tab:our_benchmark_comp}
\renewcommand{\arraystretch}{1.3}
\resizebox{\linewidth}{!}{
\begin{tabular}{lcccccccccc}
\toprule
 & Explore & Mine & Hidden & Seal & Spring & Drawer & Flash & Scroll & Placement & Avg. \\
 & Cube & Sweeper & Pattern & Checking & Compression & Match & Beacon & Discovery & Imitation & \\
\midrule
$\pi_{0.5}$ & 0\% & 1\% & 24\% & 0\% & 0\% & 21\% & 0\% & 16\% & 1\% & 7.0\% \\
Mem-0 & 0\% & 0\% & \textbf{34\%} & 0\% & 25\% & \textbf{58\%} & 0\% & 17\% & 2\% & 15.1\% \\
EventVLA & \textbf{2\%} & 1\% & 24\% & 2\% & 34\% & 51\% & 15\% & \textbf{29\%} & 6\% & 18.2\% \\
NativeMEM & 0\% & 0\% & 25\% & 0\% & 44\% & 44\% & 9\% & 19\% & 1\% & 15.8\% \\
\rowcolor{gray!20}
\textbf{Ours} & \textbf{2\%} & \textbf{93\%} & 32\% & \textbf{64\%} & \textbf{52\%} & 47\% & \textbf{38\%} & 22\% & \textbf{99\%} & \textbf{49.9\%} \\
\bottomrule
\end{tabular}}
\end{table*}

\textbf{\method substantially improves closed-loop success.}
As shown in Table~\ref{tab:our_benchmark_comp}, \method achieves 49.9\% average success, 31.7\% above the strongest baseline. It performs best on five tasks and ties for best on one more.
The largest gains occur on Placement Imitation, Mine Sweeper, and Seal Checking, which require retaining precise visual evidence after it disappears from view.\looseness=-1

\textbf{Memory-free execution struggles on these tasks.}
$\pi_{0.5}$ reaches only 7.0\% average success, and scores 0\% on four of the nine tasks.
On Mine Sweeper, we observe that $\pi_{0.5}$ often places a block inside the candidate region, but not at the location specified by the earlier visual cue.
This suggests that learning how to pick and place a block is not enough. The policy must also remember where to place it.\looseness=-1

\textbf{Texture matching remains difficult for the executor.}
\method gains most on tasks requiring appearance--location bindings.
In contrast, \method does not outperform the strongest baseline on the Appearance tasks.
To understand this, we run a diagnosis on Hidden Pattern.
Linear probes of the executor's final-layer features predict ``what the texture is'' with 99\% accuracy, but predict the matching box with only 40\% accuracy, just slightly higher than the 33\% level of random guessing.
The features retain information about each texture, but do not capture well which textures match.
Appendix~\ref{app:texture-probing} gives the setup and additional readouts.
These results suggest that the executor struggles to match textures across memory images, even when the relevant evidence is available.\looseness=-1

\subsection{Memory Selection Transfers to Unseen Visual Appearances}

We test whether the memorizer continues to select relevant frames when the scene appearance changes.
We train it on the standard appearance and apply it directly to visual OOD demonstrations.
We then train an executor on target-appearance action demonstrations using these predicted keyframes, without new memory annotations.
For Mem-0 and EventVLA, we likewise use source-trained models to generate target-domain memory annotations for action training (Table~\ref{tab:our_benchmark_OOD_comp}).
NativeMEM has no separate memory module, so we retrain it on the target-appearance data.\looseness=-1

\begin{table*}[!t]
\centering
\definecolor{memoryFreeHeader}{HTML}{E8F3E8}
\definecolor{memoryAnnotatedHeader}{HTML}{E6EFF8}
\caption{\textbf{\benchmark with unseen appearances.} Methods are grouped by whether training requires explicit memory annotations. For Mem-0, EventVLA, and \method, we reuse memory-related components trained on the standard appearance to annotate target-appearance demonstrations. We then train the action components on these demonstrations and generated annotations, without new human memory annotations. NativeMEM is retrained end-to-end on target-appearance data.}
\label{tab:our_benchmark_OOD_comp}
\renewcommand{\arraystretch}{1.3}
\resizebox{\linewidth}{!}{
\begin{tabular}{lcccccccccc}
\toprule
 & Explore & Mine & Hidden & Seal & Spring & Drawer & Flash & Scroll & Placement & Avg. \\
 & Cube & Sweeper & Pattern & Checking & Compression & Match & Beacon & Discovery & Imitation & \\
\specialrule{\lightrulewidth}{\aboverulesep}{0pt}
\rowcolor{memoryFreeHeader}
\multicolumn{11}{l}{\textit{No memory annotations required}} \\
NativeMEM & 0\% & 0\% & 31\% & 0\% & 31\% & 49\% & 8\% & 25\% & 3\% & 16.3\% \\
\specialrule{\lightrulewidth}{\aboverulesep}{0pt}
\rowcolor{memoryAnnotatedHeader}
\multicolumn{11}{l}{\textit{Requires memory annotations}} \\
Mem-0 & 0\% & 0\% & 24\% & 0\% & 28\% & 21\% & 0\% & 1\% & 0\% & 8.2\% \\
EventVLA & 0\% & 1\% & \textbf{40\%} & 0\% & 30\% & 48\% & 6\% & 4\% & 4\% & 14.8\% \\
\rowcolor{gray!20}
\textbf{Ours} & \textbf{12\%} & \textbf{79\%} & 35\% & \textbf{55\%} & \textbf{79\%} & \textbf{51\%} & \textbf{59\%} & \textbf{28\%} & \textbf{99\%} & \textbf{55.2\%} \\
\bottomrule
\end{tabular}}
\end{table*}

\textbf{Memory selection transfers to unseen appearances without new annotations.}
As shown in Table~\ref{tab:our_benchmark_OOD_comp}, \method reaches 55.2\% average success under appearance changes, while the strongest baseline reaches 16.3\%.
It achieves the best performance on eight of the nine tasks.\looseness=-1

In the visual OOD setting, the 55.2\% average success is higher than the 49.9\% under the standard appearance.
This may seem counterintuitive, but only the memorizer runs inference on an unseen appearance in this setting.
The executor is still trained on target-appearance action demonstrations, so this setting is not OOD for the executor.
On recorded demonstrations, the memorizer selects at least one frame in 95.8\% of the annotated key segments under the standard appearance, and still in 87.2\% under unseen appearances (Appendix~\ref{app:keyframe-quality}).
We can see that its selections remain useful for execution under unseen appearances.\looseness=-1

\begin{table*}[!t]
\centering
\caption{\textbf{Performance comparison on RMBench.} For Mem-0 and EventVLA, we directly use the results reported in their original papers~\citep{rmbench,eventvla}. We train NativeMEM ourselves. The reported Mem-0 results are obtained by training a separate model for each task, while the other methods are trained jointly across all tasks.}
\label{tab:rmbench_comp}
\renewcommand{\arraystretch}{1.3}
\resizebox{\linewidth}{!}{
\begin{tabular}{lcccccccccc}
\toprule
 & Observe & Rearrange & Put Back & Swap & Swap & Battery & Blocks & Cover & Press & Avg. \\
 & and Pick Up & Blocks & Block & Blocks & T & Try & Ranking Try & Blocks & Button & \\
\midrule
$\pi_{0.5}$ & 9\% & 13\% & 11\% & 24\% & 15\% & 16\% & 6\% & 0\% & 0\% & 10.4\% \\
Mem-0 & 4\% & 89\% & 90\% & 67\% & 14\% & 28\% & 18\% & 68\% & 0\% & 42.0\% \\
EventVLA & 21\% & \textbf{96\%} & 95\% & 96\% & 87\% & 35\% & \textbf{81\%} & \textbf{97\%} & 3\% & 67.9\% \\
NativeMEM & 0\% & 3\% & 0\% & 62\% & 0\% & 13\% & 0\% & 0\% & 0\% & 8.7\% \\
\rowcolor{gray!20}
\textbf{Ours} & \textbf{82\%} & 78\% & \textbf{100\%} & \textbf{99\%} & \textbf{97\%} & \textbf{56\%} & 22\% & 84\% & \textbf{10\%} & \textbf{69.8\%} \\
\bottomrule
\end{tabular}}
\end{table*}

\begin{table}[!t]
\centering
\caption{\textbf{Real-world evaluation of \method.} We report successful trials out of 30 per task.}
\label{tab:real_world}
\small
\setlength{\tabcolsep}{4pt}
\renewcommand{\arraystretch}{1.15}
\begin{tabular*}{\linewidth}{@{}l@{\extracolsep{\fill}}rrr@{}}
\toprule
Task & $\pi_{0.5}$ & EventVLA & \method \\
\midrule
Placement Imitation & 3/30 & 6/30 & \textbf{14/30} \\
Explore Cube & 11/30 & 10/30 & \textbf{17/30} \\
\bottomrule
\end{tabular*}
\end{table}

\begin{table*}[!t]
\centering
\caption{
\textbf{Ablation of memorizer training on \benchmark.} 
We report closed-loop success rates under the standard appearance. Point Annotation replaces each key segment with one fixed positive frame. The w/o SS variant removes scheduled sampling and uses only teacher-forced memory histories. Full \method uses both segment-level annotations and scheduled sampling.}
\label{tab:ablation}
\small
\renewcommand{\arraystretch}{1.3}
\setlength{\tabcolsep}{3pt}
\resizebox{\linewidth}{!}{
\begin{tabular}{lcccccccccc}
\toprule
 & Explore & Mine & Hidden & Seal & Spring & Drawer & Flash & Scroll & Placement & Avg. \\
 & Cube & Sweeper & Pattern & Checking & Compression & Match & Beacon & Discovery & Imitation & \\
\midrule
Point Annotation & 0\% & 0\% & 25\% & 0\% & 30\% & 1\% & 0\% & 1\% & 5\% & 6.9\% \\
w/o SS & 0\% & 0\% & 29\% & 2\% & \textbf{87\%} & 27\% & 0\% & \textbf{23\%} & 3\% & 19.0\% \\
\rowcolor{gray!20}
Full \method & \textbf{2\%} & \textbf{93\%} & \textbf{32\%} & \textbf{64\%} & 52\% & \textbf{47\%} & \textbf{38\%} & 22\% & \textbf{99\%} & \textbf{49.9\%} \\
\bottomrule
\end{tabular}}
\end{table*}

\subsection{The Memorizer Adapts to an Unseen Task with Few Annotations}

We test whether the memorizer can adapt to Flash Beacon after training on the other eight tasks.
We adapt the memorizer with five memory-annotated Flash Beacon demonstrations.
We then use it to generate memories for all 100 target demonstrations and train the executor.\looseness=-1

\begin{figure}[t]
    \centering
    \includegraphics[width=\linewidth]{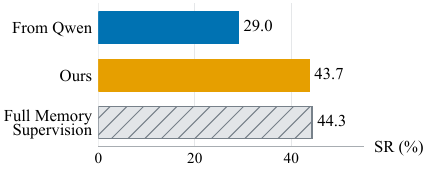}
    \caption{
    \textbf{Few-shot memorizer transfer on \benchmark.}
    After training on the other \benchmark tasks, the memorizer adapts to Flash Beacon from five annotated demonstrations. It reaches 43.7\% closed-loop success. This is close to 44.3\% with full memory supervision and above 29.0\% when adapting pretrained Qwen3-VL with the same data.
    }
    \label{fig:flash_beacon_fewshot}
\end{figure}

For comparison, the ``From Qwen'' group adapts directly from pretrained Qwen3-VL using the same five memory-annotated demonstrations, testing whether Qwen's prior knowledge alone explains the result.
The ``Full Memory Supervision'' group trains on Flash Beacon with memory annotations for all of the 100 demonstrations, serving as a positive control.
All settings use the same 100 action demonstrations and differ only in memory supervision.\looseness=-1

Training the memorizer on the source tasks improves few-shot adaptation to Flash Beacon.
With the same five annotated demonstrations, \method reaches 43.7\% success, 14.7 percentage points above the ``From Qwen'' group at 29.0\% (Figure~\ref{fig:flash_beacon_fewshot}).
This is close to the 44.3\% achieved with Full Memory Supervision.
Appendix~\ref{app:few-shot-transfer} provides the full protocol and baseline comparison.\looseness=-1

\begin{figure}[t]
\centering
\includegraphics[width=\linewidth]{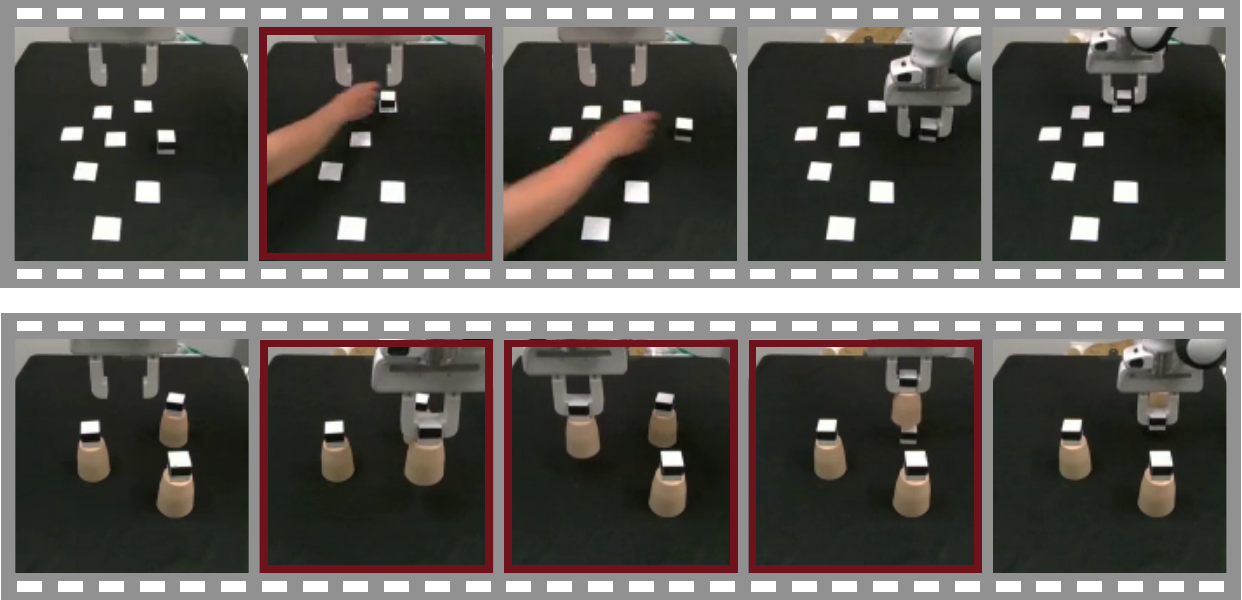}
\caption{\textbf{Real-world experiments.}
\textbf{Top:} Similar to Placement Imitation. A human places the cube at a target and then moves it back. The robot must place the cube at that target.
\textbf{Bottom:} Similar to Explore Cube. The robot must find the cube under one of three cups without revisiting any cup.}
\label{fig:real_world}
\end{figure}

\subsection{\method Is Applicable Across Benchmarks and Real-World Tasks}

\textbf{\method remains competitive on an existing memory benchmark.}
We evaluate \method on RMBench~\citep{rmbench} without changing its architecture.
As shown in Table~\ref{tab:rmbench_comp}, \method reaches 69.8\% average success, 1.9 percentage points above EventVLA at 67.9\%.
It obtains the best result on six of the nine tasks, including a 61\% gain over the strongest baseline on Observe and Pick Up.
The architecture thus remains effective on memory tasks beyond \benchmark.\looseness=-1

\textbf{\method also improves task success on a real robot.}
We evaluate \method on a Franka Panda robot using real-world counterparts of Placement Imitation and Explore Cube, as illustrated in Figure~\ref{fig:real_world}.
We collect 51 and 54 demonstrations respectively, and manually annotate their key segments.
As shown in Table~\ref{tab:real_world}, \method completes 17 more episodes than $\pi_{0.5}$ and 15 more than EventVLA across the 60 trials.
The larger gain occurs on Placement Imitation, where the robot must remember the demonstrated location after the cube is moved away.\looseness=-1

\subsection{Ablation Studies}

\textbf{Segment supervision allows multiple valid frames per event.}
We test whether the memorizer benefits from learning to capture an event segment rather than one exact frame.
We replace each annotated segment with a single positive frame, following the point annotation used by prior work.
As shown in Table~\ref{tab:ablation}, this change reduces average success by 43.0 percentage points.
Point supervision treats neighboring frames as negatives simply because they are not the selected anchor frame.
However, nearby frames may show the same evidence.
Labeling them as negative penalizes valid selections. Segment supervision avoids this problem.\looseness=-1

\textbf{Scheduled sampling improves success on multi-event tasks.}
We test whether the memorizer needs to learn from its own imperfect memory histories.
We remove scheduled sampling and train it only with teacher-forced memory histories.
As shown in Table~\ref{tab:ablation}, average success drops by 30.9 percentage points.
An early mistake changes the memory context used for every later decision.
Training with the memorizer's own predictions helps it handle earlier selection errors, especially in tasks with multiple memory events.\looseness=-1

\subsection{Discussion on Memory Capability}

\textbf{The memorizer can capture most key segments.}
Our method adopts a hierarchical architecture. This design lets us evaluate memory selection separately from execution. 
We use the simulator to verify whether the memorizer selects a frame in the key moment segment. 
On the validation set, it captures 95.8\% of key segments, while it avoids selecting a frame outside the key moment segment in 84.4\% of demonstrations. 
These results validate the training of our memorizer and show how memory selection can be evaluated directly.
Appendix~\ref{app:keyframe-quality} gives the full setup and per-task results.\looseness=-1

\textbf{Our memorizer works better than a stronger frozen VLM.}
Our memorizer is trained with the key segment annotations from humans or the simulator.
A natural question is whether a stronger VLM can select keyframes without post-training, as in BPP~\citep{bpp}.
To test this hypothesis, we follow BPP and exploit a frozen GPT-6-Astra to identify when a key event occurs.
However, querying such a VLM is costly, and thus we do not run the closed-loop evaluation.
Instead, we compare the keyframe selection on the same validation split using the same memory metrics.
It turns out that GPT-6-Astra captures only 69.5\% of key segments. 
The results demonstrate that our memorizer works clearly better than the stronger frozen VLM, which we attribute to memory training and annotations.
Therefore, general visual understanding is not strong enough for memory capability and memory modeling deserves more attention from the community.
Appendix~\ref{app:alternative-selectors} provides the full setup and per-task results.\looseness=-1

\section{Conclusion}
\label{sec:conclusion}

We study visual moment memory, which requires robots to retain precise visual details for later execution.
We introduce \benchmark, whose tasks require remembering briefly visible locations, appearances, and their associations.
We propose \method, which decouples memory from execution: a VLM memorizer stores sparse keyframes as raw images, and an executor predicts actions from them.
\method improves task success on \benchmark and a real robot, and remains competitive on RMBench.
Its memorizer transfers zero-shot to unseen appearances and adapts to a held-out task with a few memory annotations.
Future work can investigate whether broader memory training enables transfer to new tasks without memory annotations.\looseness=-1

\textbf {Limitations.}
\method runs a VLM memorizer, which adds inference cost compared with a single-model policy.
On RMBench, \method improves average success by 1.9 percentage points over the strongest evaluated baseline, but gains vary across tasks.
The executor still struggles to use retained texture information for target selection, even when the relevant evidence is available (Appendix~\ref{app:counterfactual-keyframes}).
\benchmark also uses a single embodiment and a fixed camera configuration, so the effect of viewpoint changes remains untested.\looseness=-1

\label{sec:main-text-end}

{\small
\bibliographystyle{ieeenat_fullname}
\bibliography{reference}
}

\onecolumn
\typeout{CVPR MAIN PDF PAGES: \number\numexpr\value{page}-1\relax}
\setcounter{page}{1}
\renewcommand*{\thepage}{S\arabic{page}}
\appendix
\raggedbottom
\begin{center}
  \Large\textbf{\thetitle}\par
  \vspace{0.5em}
  Supplementary Material
\end{center}

\section{Task-Relevant Visual Evidence}
\label{app:benchmark}

Table~\ref{tab:app-task-evidence} summarizes the evidence that each task in \benchmark requires the robot to retain and use after it becomes unavailable. Each key segment marks a period when one piece of evidence is visible. We annotate the key segments with privileged simulator information in place of human annotation.

\begin{table}[H]
\centering
\small
\caption{\textbf{Visual evidence and its later use in the nine tasks of \benchmark.} The final column gives the number of key segments per~demonstration.\looseness=-1}
\label{tab:app-task-evidence}
\setlength{\tabcolsep}{4pt}
\renewcommand{\arraystretch}{1.2}
\rowcolors{2}{gray!6}{white}
\begin{tabular}{@{}
>{\raggedright\arraybackslash}m{\dimexpr0.20\linewidth-6pt\relax}
>{\raggedright\arraybackslash}m{\dimexpr0.34\linewidth-6pt\relax}
>{\raggedright\arraybackslash}m{\dimexpr0.35\linewidth-6pt\relax}
>{\centering\arraybackslash}m{\dimexpr0.11\linewidth-6pt\relax}@{}}
\toprule
Task & Evidence to remember & Later use & Segments \\
\midrule
Explore Cube & Contents beneath the three covers & Retrieve the hidden cube & 3 \\
Mine Sweeper & Detector colors and their locations & Place blocks at matching locations & 2 \\
Hidden Pattern & Box--pattern associations and a target pattern & Select the matching box & 2 \\
Seal Checking & Colors revealed beneath the seals & Place blocks at the corresponding locations & 2 \\
Spring Compression & A demonstrated compression position & Reproduce the compression & 1 \\
Drawer Match & Patterns inside the three drawers & Reopen the matching drawer & 3 \\
Flash Beacon & Two transient target locations & Move the cube to both locations in order & 2 \\
Scroll Discovery & Patterns revealed by unrolling the scrolls & Stack the matching scrolls & 3 \\
Placement Imitation & Demonstrated object--position associations & Reproduce the placements & 2 \\
\bottomrule
\end{tabular}
\end{table}

\FloatBarrier
\section{Task Instructions}
\label{app:task-instructions}

The following boxes list the task instructions for the nine tasks in \benchmark. The instructions are reproduced verbatim from the environment definitions and describe the task procedure and objective. The instance-specific patterns, colors, and target locations must be inferred from visual observations. The memorizer and executor receive the task instruction. The separate memory-instruction field stored in the dataset is not supplied to the memorizer in our experiments.

\begingroup
\newtcolorbox{taskinstructionbox}[1]{
    colback=gray!4,
    colframe=black!55,
    boxrule=0.4pt,
    arc=1pt,
    left=6pt,
    right=6pt,
    top=6pt,
    bottom=6pt,
    before skip=10pt,
    after skip=10pt,
    title={#1},
    fonttitle=\bfseries,
    fontupper=\small\ttfamily,
    before upper=\raggedright
}

\begin{taskinstructionbox}{Spring Compression}
A white cube sits on a long horizontal rail connected to a spring. Before the robot moves, the cube slides to a compressed position, pauses, and returns to its starting position. As soon as it returns, close the gripper and push the cube at a steady speed to the same compressed position, then lift the closed gripper while the cube stays~there.
\end{taskinstructionbox}

\begin{taskinstructionbox}{Flash Beacon}
A white cube and two beacons are on the table. Lift the cube. When the first beacon appears, start moving toward it at a moderate speed so that it disappears before you arrive. Keep moving without returning the arm to its starting pose, and place the cube at that spot. The second beacon appears as soon as the cube touches the table. Keep holding the cube and continue directly toward the second beacon, then place and release the cube at that spot.
\end{taskinstructionbox}

\begin{taskinstructionbox}{Explore Cube}
Three identical opaque cups are spread across the table, and one cup hides a green cube. Inspect all three cups in any order. Each time, lift a cup high enough to see under it, then put it back upside down in the same place before checking the next. Once all three cups have been checked, move the cup hiding the cube aside, then grasp and lift the cube. Leave the other cups in place and leave every cup upside down on the table.
\end{taskinstructionbox}

\begin{taskinstructionbox}{Hidden Pattern}
Three boxes open together to show three different patterns, then close. A target card appears briefly with one of those patterns and then disappears. Then push forward the box that showed the same pattern as the target card.
\end{taskinstructionbox}

\begin{taskinstructionbox}{Drawer Match}
A three-level cabinet hides a pattern in each drawer. Press the upper, middle, and bottom drawer buttons in that order; each drawer opens once and then closes. Then press the upper or middle button for the drawer whose pattern matches the bottom drawer. Finish when the selected drawer is fully open.
\end{taskinstructionbox}

\begin{taskinstructionbox}{Scroll Discovery}
Three narrow scrolls hide patterned strips. From left to right, pull each strip until the whole pattern is visible, then release it so it rolls back up. After the third strip rolls up, immediately grasp one of the two scrolls that showed the same pattern and stack it on top of the other, without touching the scroll with the different~pattern.
\end{taskinstructionbox}

\begin{taskinstructionbox}{Placement Imitation}
Two cubes of different colors start at separate positions on the table. One at a time, each cube moves to an unmarked target position, pauses, and returns. Once both demonstrations are complete, place each cube at the position where it paused, then release it on the table.
\end{taskinstructionbox}

\begin{taskinstructionbox}{Mine Sweeper}\spaceskip=0.40em\relax
Two parallel black lines, a white probe, and two cubes of different colors are on the table. Move the probe directly to one end of the nearer line and scan once to the other end. Move directly to the farther line and scan it once in the opposite direction. The probe briefly changes color at one point on each line. Then put the probe back, return the arm to its starting pose, and place each cube where the probe showed the same color. Release each cube on the table.
\end{taskinstructionbox}

\begin{taskinstructionbox}{Seal Checking}
Two white stamps cover two colored spots. A tray and two cubes of different colors are also on the table. Lift each stamp by its square handle, place it fully inside the tray, and watch the spot underneath fade away. After both stamps have been checked, return the arm to its starting pose. Then place each cube where the spot showed the same color and release it on the table.
\end{taskinstructionbox}

\endgroup

\FloatBarrier
\section{Dataset Construction}
\label{app:data-construction}

\subsection{Simulation and Demonstration Collection}
We construct \benchmark in ManiSkill3 with SAPIEN physics, using a Panda manipulator, an overview camera, and a wrist camera. The data-generation pipeline records both RGB streams at $224\times224$ resolution and a control frequency of 20 Hz. Each task implements a scripted, closed-loop expert that computes actions from the simulator state and executes them through the robot controller. Demonstrations therefore contain the actions actually applied to the environment and the resulting observations.\looseness=-1

Each rollout starts from a seeded initialization, with the scene rebuilt at reset. Task-specific randomization varies object positions, target locations, and the associations between visual cues and their required responses. For example, Hidden Pattern samples distinct patterns and assigns them to boxes. Drawer Match and Scroll Discovery randomize the matching patterns. Placement Imitation, Mine Sweeper, and Seal Checking randomize color--location associations. Color pairs are sampled without replacement from a palette of eight colors, excluding pairs that are too similar under the implemented color-distance criterion. Spatial sampling also enforces task-specific separation and reachability constraints. These variations produce different memory contents while retaining the same task instruction.

The expert rolls out until success, failure, termination, or the step limit. Only successful episodes are retained. Failed attempts are discarded, and generation continues with new seeds. For tasks that expose a separate evaluation-success flag, a retained demonstration must satisfy that flag as well as the expert's success condition. Our main experiment uses 100 demonstrations per task, giving 900 demonstrations across the nine tasks.

\subsection{Automatic Memory Annotations}
Each environment records the key segments during which task-relevant evidence is available, using simulator events and task state. Each segment is stored as an inclusive pair of frame indices $[u_j,v_j]$ at the original recording rate. The metadata also records the event identity, the information to retain, and the corresponding timestamps. These segments mark valid ranges rather than fixed representative frames.

Each demonstration has one to three annotated segments, as summarized in Table~\ref{tab:app-task-evidence}. For Seal Checking, each segment ends before the revealed color becomes too faint to identify. Privileged event metadata supplies supervision for memory selection and is not included in the memorizer's or executor's observation input.

\subsection{Observation and Action Alignment}
Raw trajectories are saved in HDF5, with accompanying JSON metadata containing the episode seed, task instruction, phase intervals, and memory annotations. A rollout with $T$ actions contains observations from $o_0$ through $o_T$. Conversion creates $T$ training records, pairing each action $a_t$ with its preceding observation $o_t$ and timestamp. The terminal observation is not assigned an additional action. Both camera streams and the key segment boundaries use this frame indexing.

Hidden Pattern, Spring Compression, and Placement Imitation begin with an observation phase in which the environment presents a visual demonstration without requesting policy actions. These observations remain available for memory selection. For \method, the recorded stationary actions during these observation phases are excluded from action supervision, and the executor is trained only on the execution phase. The remaining tasks require robot actions to acquire the evidence, so their recorded actions remain eligible for supervision throughout the trajectory.

Our EventVLA implementation instead retains supervision on these stationary actions, because its keyframe predictions are derived from action hidden states. It therefore learns these actions even though the environment does not request them during observation.\looseness=-1

\subsection{Training-Data Conversion}
We convert the recordings into LeRobot-format datasets with per-episode videos, tabular state--action records, task metadata, and a separate file of key-segment annotations. The converter checks camera streams, temporal alignment, and segment bounds, and rejects missing or malformed segments. It reconstructs the controller's commanded targets from recorded simulator states to express end-effector actions in the overview-camera coordinate system. For the executor used in this paper, actions comprise relative end-effector translation, relative rotation represented in Rot6D, and a gripper command. Dataset conversion preserves the annotated key segments. Anchor sampling is performed during training, as described in Section~\ref{sec:methodology}.

\subsection{Visual OOD Data}
Visual OOD environments use the same task implementations and expert procedures, with altered room and table appearance, lighting, and additional background or tabletop distractors outside the manipulation area. Appearance randomization uses a separate seed-derived random-number stream, preserving the task randomization for a given seed. Task mechanics, memory events, and success conditions are shared with the standard setting, and OOD trajectories pass through the same recording and conversion pipeline. In the appearance-transfer experiment, target-domain executor training uses keyframes predicted by the transferred memorizer rather than the simulator's target-domain memory annotations.

\FloatBarrier
\section{Implementation and Training Details}
\label{app:implementation}

We implement the memorizer with Qwen3-VL-4B~\citep{qwen3vl} and the executor with $\pi_{0.5}$~\citep{pi05}. On \benchmark, the memorizer receives only overview images. The executor receives current overview and wrist images, while its keyframe stack contains only overview images. The memorizer receives neither privileged memory instructions nor absolute historical frame indices. \benchmark operates at 20 Hz, and the memorizer samples observations at 5 Hz. Table~\ref{tab:app-config} lists the standard-appearance configuration.\looseness=-1

For the main experiments on \benchmark and RMBench, we reserve 20\% of the demonstrations with key segment annotations for memorizer validation and use the remaining 80\% for training. On \benchmark, this gives the memorizer 80 training and 20 validation demonstrations per task. The executor uses all 100 demonstrations. We select the memorizer checkpoint with the highest mean of segment accuracy and non-keyframe accuracy on the validation set. These metrics are defined in Appendix~\ref{app:keyframe-quality}.

\begin{table}[H]
\centering
\small
\caption{\textbf{Configuration of \method on \benchmark under the standard appearance.}}
\label{tab:app-config}
\begin{tabular}{@{}ll@{}}
\toprule
Setting & Value \\
\midrule
Control frequency & 20 Hz \\
Maximum memorizer training steps & 10,000 \\
Selected memorizer checkpoint & Step 10,000 \\
Memorizer sampling rate (validation/inference) & 5 Hz \\
Recent observation buffer & 10 images \\
Positive decision threshold & 0.5 \\
Maximum model-feedback probability & 0.4 \\
Memorizer focal loss $(\alpha, \gamma)$ & $(0.8, 2.5)$ \\
Executor training steps & 60,000 \\
Executor action chunk size (training) & 10 actions \\
Executor action chunk size (execution) & 10 actions \\
\bottomrule
\end{tabular}
\end{table}

Table~\ref{tab:app-rmbench-config} lists the RMBench configuration. Actions are executed at 16.67 Hz. For memorizer validation and inference, we select one frame out of every three, giving a sampling rate of 5.56 Hz. Training uses temporal sampling augmentation, with frame intervals of 1, 2, or 3 sampled with probabilities 0.1, 0.1, and 0.8, respectively. During evaluation, the executor executes the first 40 actions of each predicted chunk before predicting the next chunk. For the RMBench tasks, we manually annotate key segments for \method. We do not use the subtask annotations provided by RMBench.

\begin{table}[H]
\centering
\small
\caption{\textbf{Configuration of \method on RMBench.} The memorizer checkpoint is selected using the validation criterion above.}
\label{tab:app-rmbench-config}
\begin{tabular}{@{}ll@{}}
\toprule
Setting & Value \\
\midrule
Control frequency & 16.67 Hz \\
Maximum memorizer training steps & 20,000 \\
Selected memorizer checkpoint & Step 18,000 \\
Memorizer sampling rate (validation/inference) & 5.56 Hz \\
Recent observation buffer & 10 images \\
Positive decision threshold & 0.5 \\
Maximum model-feedback probability & 0.9 \\
Memorizer focal loss $(\alpha, \gamma)$ & $(0.8, 2.5)$ \\
Executor training steps & 90,000 \\
Executor action chunk size (training) & 50 actions \\
Executor action chunk size (execution) & 40 actions \\
\bottomrule
\end{tabular}
\end{table}

The anchor labels described in Section~\ref{sec:identifier-training} and Figure~\ref{fig:method}(b) assume a teacher-forced history in which the anchor is stored. Scheduled sampling instead requires labels to reflect the memory history shown to the memorizer. Within a key segment, frames before the sampled anchor are masked from the loss until a representative is stored. From the anchor onward, the target is Yes until the visible history contains a representative of that segment. It then becomes No. Frames outside all segments receive No. For example, if the memorizer misses an anchor, a later frame in the same segment remains a positive example. This history-dependent supervision allows recovery from missed selections.

During training, the executor retains only keyframes selected by the trained memorizer before the current frame and masks unused memory slots. On \benchmark, actions follow the representation in Appendix~\ref{app:data-construction}. State and action normalization use training-data statistics. On RMBench, the executor instead uses dual-arm absolute joint-position and gripper actions.

For executor training, each frame selected offline by the memorizer is stored as a single-frame key segment. The executor training pipeline then uses these predicted selections in the same way as annotated segments. We use this procedure in both the main and transfer experiments. The images, actions, and task instructions in the data are not modified.

\FloatBarrier
\section{Keyframe Selection Quality}
\label{app:keyframe-quality}

When a task fails, the outcome alone cannot tell us whether the memorizer missed a visual cue.
We check which cues it captures, how often it saves the same cue more than once, and whether it saves unrelated frames.

On \benchmark, the 20\% validation split contains 20 demonstrations per task, or 180 in total. This is the checkpoint-selection set, not an independent test set. Evaluation replays each demonstration autoregressively, so the memorizer's previous decisions determine its later memory context.

Segment hit rate measures the fraction of key segments containing at least one selected frame. Segment accuracy requires exactly one selection per segment. To measure selections outside the key segments, we also report the fraction of demonstrations with no such selection, termed non-keyframe accuracy. Any such selection stays in memory for the rest of the demonstration, so this metric reflects how often the memory remains free of irrelevant frames.

\begin{table}[H]
\centering
\small
\caption{\textbf{Memorizer validation on \benchmark under the standard appearance.} Hit rate and segment accuracy are computed over key segments. Non-keyframe accuracy is computed over demonstrations. The pooled row aggregates counts across tasks.}
\label{tab:app-selector}
\setlength{\tabcolsep}{4pt}
\renewcommand{\arraystretch}{1.15}
\begin{tabular*}{\linewidth}{@{}l@{\extracolsep{\fill}}ccc@{}}
\toprule
Task & Segment hit rate (\%) & Segment accuracy (\%) & Non-keyframe accuracy (\%) \\
\midrule
Explore Cube & 98.3 & 96.7 & 90.0 \\
Mine Sweeper & 100.0 & 100.0 & 100.0 \\
Hidden Pattern & 100.0 & 100.0 & 100.0 \\
Seal Checking & 97.5 & 95.0 & 80.0 \\
Spring Compression & 100.0 & 90.0 & 100.0 \\
Drawer Match & 100.0 & 100.0 & 100.0 \\
Flash Beacon & 97.5 & 87.5 & 80.0 \\
Scroll Discovery & 86.7 & 83.3 & 65.0 \\
Placement Imitation & 85.0 & 82.5 & 45.0 \\
\midrule
\textbf{Pooled} & \textbf{95.8} & \textbf{93.0} & \textbf{84.4} \\
\bottomrule
\end{tabular*}
\end{table}

As shown in Table~\ref{tab:app-selector}, the memorizer selects at least one frame in 383 of 400 key segments and exactly one in 372. It makes no selection outside the key segments in 152 of 180 demonstrations.

\Needspace{6\baselineskip}
We also evaluate the same memorizer on unseen appearances without further fine-tuning. Table~\ref{tab:app-selector-ood} reports autoregressive evaluation on all 100 demonstrations per task, or 900 in total, with the decision threshold fixed at 0.5. Unlike the standard-appearance validation split, these demonstrations are not used to select the memorizer checkpoint.

\begin{table}[H]
\centering
\small
\caption{\textbf{Memorizer validation on \benchmark under unseen appearances.} The memorizer is unchanged from Table~\ref{tab:app-selector}. Results cover 100 demonstrations per task and use the same strict segment criteria. The pooled row aggregates counts across tasks.}
\label{tab:app-selector-ood}
\setlength{\tabcolsep}{4pt}
\renewcommand{\arraystretch}{1.15}
\begin{tabular*}{\linewidth}{@{}l@{\extracolsep{\fill}}ccc@{}}
\toprule
Task & Segment hit rate (\%) & Segment accuracy (\%) & Non-keyframe accuracy (\%) \\
\midrule
Explore Cube & 91.7 & 88.3 & 54.0 \\
Mine Sweeper & 99.5 & 99.5 & 100.0 \\
Hidden Pattern & 97.5 & 97.5 & 97.0 \\
Seal Checking & 98.0 & 98.0 & 88.0 \\
Spring Compression & 10.0 & 10.0 & 4.0 \\
Drawer Match & 95.7 & 94.3 & 89.0 \\
Flash Beacon & 90.0 & 79.0 & 61.0 \\
Scroll Discovery & 84.7 & 80.3 & 37.0 \\
Placement Imitation & 74.0 & 73.0 & 47.0 \\
\midrule
\textbf{Pooled} & \textbf{87.2} & \textbf{84.7} & \textbf{64.1} \\
\bottomrule
\end{tabular*}
\end{table}

Spring Compression has low strict segment scores, but 88 of its 90 missed demonstrations contain a selected frame only 0.10\,s before the annotated start, less than the memorizer's 0.20\,s sampling period. Across all 90 misses, the selected frame nearest the segment differs from the segment's mean spring position by less than 0.4\,mm. This is far below the task's 45\,mm compression-position tolerance for successful execution. These keyframes therefore retain the relevant compression information despite falling outside the annotated key segment, consistent with the high task success rate.

\FloatBarrier
\section{Analysis of Memory Selection, Use, and Execution}
\label{app:result-analysis}

In Table~\ref{tab:our_benchmark_comp}, \method achieves its largest gains on tasks requiring appearance--location bindings, while tasks in the Appearance category remain challenging for all evaluated methods. We study Hidden Pattern, Flash Beacon, and Seal Checking as examples of the Appearance, Spatial, and Spatial \& Appearance categories, respectively. We compare \method with EventVLA, the baseline with the highest average success in the standard setting.

We examine whether relevant evidence is retained, whether it influences subsequent actions, and whether the required physical operations are completed. We also evaluate whether keyframe selection remains effective under visual appearance changes. All analyses use the final checkpoints on ten fixed test scenes per task, without further training.

\subsection{Counterfactual Keyframe Content}
\label{app:counterfactual-keyframes}

If a stored cue guides the robot's choice, changing the cue should change which target it chooses.

\textbf{Protocol.}
We use the same scene seeds for both methods.
Each scene is assigned original and counterfactual conditions.
The counterfactual branch runs only when the intervention point is reached. It then edits the cue content in eligible keyframes selected by the method itself.
Scenes without an intervention point or an eligible selected frame remain in the counts.
For Seal Checking, we exchange the two colors of the seal in the stored keyframes.
For Flash Beacon, we move the remembered first beacon to the scene's second target location, and evaluate only the response to the first beacon.
For Hidden Pattern, we replace the remembered target-card pattern with the pattern in another box.

In Table~\ref{tab:app-content-interventions}, \textit{Reached} counts scenes that reach the intervention point, and \textit{Modified} counts scenes whose selected keyframes are changed, both out of ten scenes.
\textit{Original} reports successful completion under the original condition out of all ten scenes.
\textit{Changed} counts modified scenes in which the observed target choice changes to the target specified by the counterfactual cue.
For Seal Checking, we count scenes where at least one cube switches to its counterfactual placement target.
For Flash Beacon, we use the nearest target at first contact or first-cue completion as a spatial proxy for target choice.
For Hidden Pattern, we compare the selected boxes.
Reached scenes remain unmodified if no selected keyframe falls within the cue interval eligible for editing.\looseness=-1

\begin{table}[H]
\centering
\small
\captionsetup{skip=6pt}
\caption{\textbf{Target changes after keyframe interventions.} \textit{Reached}, \textit{Modified}, and \textit{Original} use all ten scenes as the denominator. \textit{Changed} counts switches to the counterfactual target among modified scenes, without requiring successful task completion.}
\label{tab:app-content-interventions}
\setlength{\tabcolsep}{5pt}
\renewcommand{\arraystretch}{1.15}
\begin{tabular*}{\linewidth}{@{\extracolsep{\fill}}llcccc@{}}
\toprule
\textbf{Task} & \textbf{Method} & \textbf{Reached} & \textbf{Modified} & \textbf{Original} & \textbf{Changed} \\
\midrule
\multirow{2}{*}{Seal Checking} & \method & 7/10 & 7/10 & 6/10 & \textbf{7/7} \\
 & EventVLA & 0/10 & 0/10 & 0/10 & -- \\
\addlinespace[4pt]
\multirow{2}{*}{Flash Beacon} & \method & 10/10 & 8/10 & 4/10 & \textbf{6/8} \\
 & EventVLA & 10/10 & 7/10 & 2/10 & 3/7 \\
\addlinespace[4pt]
\multirow{2}{*}{Hidden Pattern} & \method & 10/10 & 10/10 & 3/10 & 0/10 \\
 & EventVLA & 10/10 & 10/10 & 0/10 & \textbf{1/10} \\
\bottomrule
\end{tabular*}
\end{table}

\textbf{Color and spatial evidence.}
On Seal Checking, exchanging the stored colors changes \method's placements to match the new associations (Table~\ref{tab:app-content-interventions}).
Six scene pairs succeed under both their original and counterfactual conditions.
This shows that the remembered color--location bindings influence placement decisions.
EventVLA does not reach the intervention point in these runs, so we examine it separately below.

On Flash Beacon, both methods change the approached target in some pairs.
All observed \method switches follow the counterfactual cue, while EventVLA also exhibits a switch in the opposite direction.
\method additionally shows a larger shift toward the alternative target at matched post-intervention times, measured by the cube's relative distance to the two targets.
These responses indicate sensitivity to stored spatial cues, even when the subsequent manipulation fails.

\textbf{Texture matching.}
On Hidden Pattern, selected keyframes cover both the box-pattern and target-card events in all ten \method scenes and eight EventVLA scenes.
Among these scenes, \method never switches to the counterfactual target.
EventVLA does so only once, without completing the push.
The failures therefore do not come from missing frames.
Both methods struggle to use retained texture information for target selection.

\subsection{Evidence Acquisition and Physical Execution}

\textbf{Seal Checking.}
Most EventVLA runs reveal both cues but fail to return to the required home position before placement.
\method also fails at this stage in some runs.

To examine placement after successful evidence acquisition, we give both methods the same expert acquisition trajectory.
Each method selects its own keyframes and resumes control with a fresh plan after the cues disappear.
\method completes both the original and counterfactual placements in seven of the ten scene pairs.
EventVLA completes neither condition even after successful evidence acquisition.
Its failures include unsuccessful grasps, drops, and releases at incorrect locations.

\textbf{Flash Beacon.}
Some \method runs reach the correct target region but fail to satisfy the grasp requirement.
Both methods also make contacts away from the targets.
Thus, a response to the remembered cue does not always lead to successful~execution.

\Needspace{18\baselineskip}
\subsection{Memory Selection Transfer under Unseen Appearances}
\label{app:selector-transfer-analysis}

\begin{wraptable}{r}{0.5\textwidth}
\vspace{-8pt}
\centering
\small
\captionsetup{skip=6pt,justification=raggedright,singlelinecheck=false}
\caption{\textbf{Memory selection transfer under unseen appearances.} Scenes covering both cue events, out of ten per condition.}
\label{tab:app-selector-transfer}
\setlength{\tabcolsep}{5pt}
\renewcommand{\arraystretch}{1.15}
\begin{tabular*}{\linewidth}{@{\extracolsep{\fill}}lcc@{}}
\toprule
\textbf{Method} & \textbf{Standard} & \textbf{OOD} \\
\midrule
\multicolumn{3}{@{}l}{\textit{Seal Checking}} \\
\method & \textbf{8/10} & \textbf{8/10} \\
EventVLA & 6/10 & 7/10 \\
\addlinespace[4pt]
\multicolumn{3}{@{}l}{\textit{Flash Beacon}} \\
\method & \textbf{8/10} & \textbf{9/10} \\
EventVLA & 7/10 & 3/10 \\
\bottomrule
\end{tabular*}
\vspace{-6pt}
\end{wraptable}

To test whether unseen appearances disrupt memory selection, we freeze both methods and evaluate their keyframe selection on matched expert demonstrations under the standard and unseen appearances.
Actions and physical states are held fixed, and each method selects keyframes in its own way.
We measure whether the selected frames cover both annotated cue events in each scene.
This comparison evaluates memory selection independently of action execution.

\method maintains coverage under unseen appearances (Table~\ref{tab:app-selector-transfer}).
EventVLA remains effective on Seal Checking but loses coverage on Flash Beacon.

\textbf{Discussion.}
\method adjusts its actions to edited color and spatial cues, and its memorizer maintains cue coverage under unseen appearances.
EventVLA misses some cue events, responds less consistently to edited spatial cues, and also fails in evidence acquisition and manipulation.
On Hidden Pattern, both methods rarely follow the edited texture even when keyframes cover both relevant events. This suggests a shared difficulty in using retained texture information.

\FloatBarrier
\section{Probing Texture Recognition and Matching}
\label{app:texture-probing}

In Hidden Pattern, the memorizer captures both the box patterns and the target card in all ten scenes examined in Appendix~\ref{app:counterfactual-keyframes}.
Yet when we change the pattern on the stored card, the robot never switches to the box with the new pattern.
We use linear probes to test two properties of the executor's features. The first is whether they encode which texture appears in each region. The second is whether they encode which box matches the card.
We fit the probes on Hidden Pattern's 100 training demonstrations and test them on 50 independent evaluation scenes.

\begin{wraptable}{r}{0.5\textwidth}
\centering
\small
\caption{\textbf{Linear probing on Hidden Pattern.} We score one texture prediction per region, giving four predictions per test scene. Target-box prediction has three classes, with a uniform-chance accuracy of 33.3\%.\looseness=-1}
\label{tab:texture-probing}
\setlength{\tabcolsep}{2pt}
\begin{tabular*}{\linewidth}{@{}>{\raggedright\arraybackslash}p{0.49\linewidth}@{\extracolsep{\fill}}rr@{}}
\toprule
Probe & \shortstack[r]{Correct} & Accuracy \\
\midrule
Recognition: region features & 198 / 200 & 99.0\% \\
Matching: region features & 20 / 50 & 40.0\% \\
Matching: text features & 22 / 50 & 44.0\% \\
\bottomrule
\end{tabular*}
\end{wraptable}

\textbf{Features.}
Each input contains the current overview and wrist images, robot state, task instruction, and two memory images.
One memory image shows the textures on the three candidate boxes. The other shows the target texture on the query card.
We use the model's native preprocessing, prefix embeddings, and attention mask.
We extract contextualized visual tokens from the final layer of the PaliGemma language backbone, rather than from the vision encoder alone.
Each image has a $16\times16$ token grid.
Using privileged region locations, we take a $4\times4$ window near the center of each of the three candidate textures and the query texture.
We flatten each window into a feature vector. The privileged locations give the probes a location prior.\looseness=-1

\textbf{Linear probes.}
For each region, we train a separate classifier to predict which of the eight textures appears there.
For target selection, we concatenate the four region features and train a three-class classifier to predict the target box.
This probe uses only target-box labels.
As a control, we predict the target box from the mean feature across non-padding text tokens.

\textbf{Results.}
As shown in Table~\ref{tab:texture-probing}, the probes identify individual textures with 99.0\% accuracy.
The probe using the four region features selects the matching box with 40.0\% accuracy, slightly above the chance level of 33.3\%.
The control reaches 44.0\%.
Thus, the model represents individual textures but not whether two textures match.

\FloatBarrier
\section{Few-Shot Cross-Task Memory Transfer}
\label{app:few-shot-transfer}

We ask whether training on other tasks helps the memorizer learn what to save in a new task with few annotations.
Flash Beacon is the target task and is excluded from the eight source tasks.
The target task provides 100 action demonstrations, but only five provide memory annotations in the few-shot settings.
We compare source-to-target adaptation with two controls, full target memory supervision and target-only adaptation from the pretrained backbone.
All six variants of \method and EventVLA share 300 evaluation seeds.

\textbf{Training settings.}
For \method, source-to-target adaptation first trains the memorizer on the other eight tasks, then fine-tunes it using five annotated Flash Beacon demonstrations (Ours in Figure~\ref{fig:flash_beacon_fewshot}).
The adapted memorizer labels all 100 target demonstrations offline, including the five used for adaptation.
The executor is trained on these demonstrations using the predicted keyframes.

Full Memory Supervision is the positive control. We train the memorizer with all available target memory annotations, then freeze it to generate keyframes for the target demonstrations. The executor is trained on these predicted memories and demonstration actions, following the same procedure as the other settings.
Both components are trained only on Flash Beacon, without source-task training.
This setting gives the success rate under full memory supervision on Flash Beacon.

From Qwen is the target-only control.
It fine-tunes the memorizer from pretrained Qwen3-VL on the same five annotated demonstrations, without source-task training.
It then uses the same automatic-labeling and executor-training procedure as Ours.
This control tests whether few-shot success comes from the pretrained backbone alone or from memory selection learned on source tasks.

\begin{table}[htbp]
\centering
\caption{\textbf{Few-shot cross-task transfer of memory selection.} Flash Beacon success rates (\%) on 300 shared evaluation seeds. Full Memory Supervision uses all target memory annotations. Target only adapts from the pretrained backbone without source-task training. Source$\rightarrow$Target first trains on the eight source tasks. For \method, these two few-shot settings are From Qwen and Ours in Figure~\ref{fig:flash_beacon_fewshot}. Both few-shot settings use five annotated target demonstrations. All settings use 100 target action demonstrations.}
\label{tab:app-few-shot-transfer}
\small
\setlength{\tabcolsep}{3pt}
\renewcommand{\arraystretch}{1.15}
\begin{tabular*}{0.5\linewidth}{@{}l@{\extracolsep{\fill}}rr@{}}
\toprule
Memory supervision & \method & EventVLA \\
\midrule
Target only & 29.0 & 5.7 \\
Source$\rightarrow$Target & 43.7 & 5.0 \\
Full Memory Supervision & 44.3 & 7.7 \\
\bottomrule
\end{tabular*}
\end{table}

We apply the same three settings to EventVLA.
Each method keeps its own architecture, and all settings use the same target action demonstrations.

\textbf{Results.}
With five annotated target demonstrations, \method with source-to-target adaptation reaches 43.7\% success, close to 44.3\% under Full Memory Supervision (Table~\ref{tab:app-few-shot-transfer}).
From Qwen reaches 29.0\%, 14.7 percentage points lower.
The first comparison shows that few-shot transfer nearly matches full supervision. The second shows that the gain comes from memory training on the source tasks rather than from the pretrained backbone.\looseness=-1

EventVLA reaches 5.0\% with source-to-target adaptation, compared with 5.7\% for target-only adaptation and 7.7\% under full memory supervision.
EventVLA shows no transfer benefit.
Its low positive-control success also shows that EventVLA struggles on Flash Beacon even with full memory supervision.\looseness=-1\par

\FloatBarrier
\section{Real-World Experiments}
\label{app:real-world-setup}

\subsection{Experimental Setup}

We evaluate \method, $\pi_{0.5}$, and EventVLA on a Franka Panda robot in two tasks adapted from \benchmark. We keep their simulation names, Placement Imitation and Explore Cube. All methods use only a single overview camera, with no wrist camera. We collect 51 and 54 demonstrations, respectively, and manually annotate their key segments because privileged simulator information is unavailable in real-world experiments. Each method is evaluated over 30 trials per task.

In Placement Imitation, seven cards are scattered across the tabletop. A human first places a cube on one card, then moves it in front of the robot. The robot must remember this location and return the cube to the correct card within 60 seconds.

In Explore Cube, a cube is hidden under one of three inverted cups. The robot must inspect the cups, locate the cube, and grasp and lift it within 120 seconds. Revisiting a previously inspected cup or knocking over any cup counts as a failure.

For both tasks, a force-triggered protective stop also counts as a failure. We stop the robot when the Euclidean norm of the three external-force components at the end effector, estimated by the Franka controller, exceeds 100~N. This can occur when the robot presses into the tabletop or an object.

To reduce slippage, we wrap the cube in sandpaper and attach sandpaper-covered blocks to the cups as grasping surfaces. These modifications improve grasp stability. They do not change what the robot must remember, which is the target card or the previously inspected cups. All methods use these modified objects.

\subsection{Training and Inference}

Table~\ref{tab:app-real-world-config} lists the hyperparameters. Inference runs on a remote server with an NVIDIA H20 GPU. The nominal control and memorizer sampling rates are 15 Hz and 5 Hz, respectively. Observations captured while awaiting an action response are discarded, and this waiting interval does not advance the policy's time index. Network and inference delays therefore reduce the effective wall-clock rates.

The 60-second and 120-second task limits are measured in wall-clock time and include all request waiting time. Network fluctuations can therefore change the number of executed actions across trials.

\begin{table}[H]
\centering
\small
\caption{\textbf{Configuration of \method in the real-world experiments.} Frequencies refer to the logical timeline, excluding \mbox{action-request latency}.\looseness=-1}
\label{tab:app-real-world-config}
\begin{tabular}{@{}ll@{}}
\toprule
Setting & Value \\
\midrule
Control frequency & 15 Hz \\
Maximum memorizer training steps & 8,000 \\
Selected memorizer checkpoint & Step 8,000 \\
Memorizer sampling rate (validation/inference) & 5 Hz \\
Recent observation buffer & 10 images \\
Positive decision threshold & 0.5 \\
Maximum model-feedback probability & 0.8 \\
Memorizer focal loss $(\alpha, \gamma)$ & $(0.8, 2.5)$ \\
Executor training steps & 20,000 \\
Executor action chunk size (training) & 10 actions \\
Executor action chunk size (execution) & 10 actions \\
\bottomrule
\end{tabular}
\end{table}

\subsection{Failure Analysis}

In Explore Cube, $\pi_{0.5}$ repeatedly inspects the same cup or returns to a previously inspected cup. These failures are consistent with difficulty tracking exploration history without memory. We also observe incomplete inspections. The robot sometimes lifts the cup hiding the cube only slightly, then lowers it before the cube becomes visible.

Once \method sees the cube, it usually moves the covering cup aside and grasps the cube. By contrast, $\pi_{0.5}$ sometimes presses into the cube and triggers the protective stop, or closes its gripper beside the cube.

In Placement Imitation, $\pi_{0.5}$ fails to return the cube to the demonstrated location after the human moves it away. \method also makes placement errors, but achieves higher success than $\pi_{0.5}$.

EventVLA succeeds in 6 of 30 trials in Placement Imitation and 10 of 30 trials in Explore Cube. In Placement Imitation, the key moment is produced by the human demonstrator rather than by the robot. EventVLA predicts keyframes from its own predicted future actions. Events caused by an external agent are not reflected in these actions, which makes them difficult for EventVLA to select. \method instead selects keyframes from visual observations, regardless of which agent causes the event. In Explore Cube, we do not observe a failure mode specific to the memory of EventVLA. Most of its failures are failed grasps, which reflect imprecise action execution rather than incorrect keyframe selection.

\FloatBarrier
\section{Alternative Keyframe Selectors}
\label{app:alternative-selectors}

We first test whether a frozen VLM can select useful keyframes when a task-specific prompt tells it what to look for, following BPP~\citep{bpp}.
We also test whether a smaller recurrent model can learn to select keyframes from the same annotations as our~memorizer.

\subsection{BPP-Style Selection with a Frozen VLM}
\label{app:bpp-selector}

Following BPP, we prompt a frozen VLM, GPT-6-Astra, to decide whether each frame shows a key event.
A hand-written prompt specifies the key events of each task.
We use the overview camera as input, since the task-relevant information is visible in it.
The original BPP queries the VLM at 1\,Hz, which can miss key segments shorter than one second. We therefore query at 2\,Hz, and each query contains the current frame and the frame 0.5\,s earlier.
A keyframe is marked at each transition from ``no'' to ``yes''.
Because of the cost of querying the VLM, we do not run closed-loop evaluation.
Since the selected keyframes directly determine the executor's input, we instead evaluate keyframe selection on the same validation split and with the same metrics as Table~\ref{tab:app-selector}.

\begin{table}[t]
\centering
\small
\caption{\textbf{BPP-style selection with a frozen VLM.} Keyframe selection on the \benchmark validation split under the standard appearance, with 20 demonstrations per task. Metrics follow Table~\ref{tab:app-selector}. The pooled row aggregates counts across tasks.}
\label{tab:app-bpp}
\setlength{\tabcolsep}{4pt}
\renewcommand{\arraystretch}{1.15}
\begin{tabular*}{\linewidth}{@{}l@{\extracolsep{\fill}}cccccc@{}}
\toprule
 & \multicolumn{2}{c}{Segment hit rate (\%)} & \multicolumn{2}{c}{Segment accuracy (\%)} & \multicolumn{2}{c}{Non-keyframe accuracy (\%)} \\
\cmidrule(lr){2-3} \cmidrule(lr){4-5} \cmidrule(lr){6-7}
Task & BPP & \method & BPP & \method & BPP & \method \\
\midrule
Explore Cube & \textbf{100.0} & 98.3 & \textbf{100.0} & 96.7 & 0.0 & \textbf{90.0} \\
Mine Sweeper & 72.5 & \textbf{100.0} & 67.5 & \textbf{100.0} & 0.0 & \textbf{100.0} \\
Hidden Pattern & \textbf{100.0} & \textbf{100.0} & 85.0 & \textbf{100.0} & 55.0 & \textbf{100.0} \\
Seal Checking & \textbf{100.0} & 97.5 & \textbf{100.0} & 95.0 & 10.0 & \textbf{80.0} \\
Spring Compression & 0.0 & \textbf{100.0} & 0.0 & \textbf{90.0} & 35.0 & \textbf{100.0} \\
Drawer Match & \textbf{100.0} & \textbf{100.0} & \textbf{100.0} & \textbf{100.0} & 0.0 & \textbf{100.0} \\
Flash Beacon & \textbf{100.0} & 97.5 & \textbf{100.0} & 87.5 & \textbf{100.0} & 80.0 \\
Scroll Discovery & 15.0 & \textbf{86.7} & 15.0 & \textbf{83.3} & \textbf{90.0} & 65.0 \\
Placement Imitation & 0.0 & \textbf{85.0} & 0.0 & \textbf{82.5} & 0.0 & \textbf{45.0} \\
\midrule
\textbf{Pooled} & 69.5 & \textbf{95.8} & 67.5 & \textbf{93.0} & 32.2 & \textbf{84.4} \\
\bottomrule
\end{tabular*}
\end{table}

As shown in Table~\ref{tab:app-bpp}, the frozen VLM captures 69.5\% of key segments, compared with 95.8\% for \method.
It fails on Spring Compression and Placement Imitation and is weak on Scroll Discovery.
BPP marks about 4.2\% of the non-key frames it evaluates as keyframes, compared with 0.14\% for \method.
Over a long demonstration, BPP therefore selects more false keyframes than true key events.
We identify two causes.
First, each query sees only two images. Spring Compression and Placement Imitation require the moment when an object reaches its farthest position, but two frames cannot show whether it has started to return.
Second, every transition from ``no'' to ``yes'' creates a keyframe. After a correct detection, any later error creates a new transition and thus a false keyframe, so errors accumulate over long demonstrations.
\method mitigates both problems in training. Anchor sampling treats any frame in a key segment as a valid choice, so the memorizer can still select a later frame if it misses an earlier one. Scheduled sampling trains the memorizer on histories that include its own predictions (Section~\ref{sec:keyframe-segments}).

\subsection{Learned Recurrent Selector}
\label{app:rnn-selector}

To separate the effect of keyframe selection from that of storing raw images, we replace the Qwen3-VL-4B memorizer with a learned recurrent selector and keep the memory format unchanged.
The selector encodes each frame and the task instruction with CLIP, and a two-layer GRU predicts whether each frame is a keyframe.
It is trained with the same key segment annotations, anchor sampling, and scheduled sampling as the memorizer.
The selected frames are given to the executor as raw images.
The executor uses the same $\pi_{0.5}$ architecture as \method and is retrained on the same demonstrations with the selected keyframes.
Under unseen appearances, we follow the same protocol as \method in Table~\ref{tab:our_benchmark_OOD_comp}. The selector trained on the standard appearance annotates the target-appearance demonstrations, and the executor is trained on these demonstrations and the selected keyframes.

\begin{table}[H]
\centering
\caption{\textbf{Learned recurrent selector.} Success rates on \benchmark under the standard and unseen appearances. Only the keyframe selector differs. The executor architecture, training data, and memory format are the same.}
\label{tab:app-rnn-selector}
\renewcommand{\arraystretch}{1.3}
\resizebox{\linewidth}{!}{
\begin{tabular}{llcccccccccc}
\toprule
 & & Explore & Mine & Hidden & Seal & Spring & Drawer & Flash & Scroll & Placement & Avg. \\
Appearance & Selector & Cube & Sweeper & Pattern & Checking & Compression & Match & Beacon & Discovery & Imitation & \\
\midrule
Standard & RNN & 0\% & 3\% & \textbf{32\%} & 2\% & 48\% & 42\% & 6\% & \textbf{25\%} & 6\% & 18.2\% \\
Standard & \method & \textbf{2\%} & \textbf{93\%} & \textbf{32\%} & \textbf{64\%} & \textbf{52\%} & \textbf{47\%} & \textbf{38\%} & 22\% & \textbf{99\%} & \textbf{49.9\%} \\
\midrule
Unseen & RNN & 0\% & 0\% & 25\% & 0\% & 36\% & 27\% & 0\% & 26\% & 1\% & 12.8\% \\
Unseen & \method & \textbf{12\%} & \textbf{79\%} & \textbf{35\%} & \textbf{55\%} & \textbf{79\%} & \textbf{51\%} & \textbf{59\%} & \textbf{28\%} & \textbf{99\%} & \textbf{55.2\%} \\
\bottomrule
\end{tabular}}
\end{table}

As shown in Table~\ref{tab:app-rnn-selector}, the average success drops sharply with the recurrent selector.
The drop is largest on tasks where the evidence appears only briefly, such as Mine Sweeper and Seal Checking. Once the selector misses that moment, the memory contains no frame with this evidence.
On the Appearance tasks, the two selectors are close. This is consistent with Appendix~\ref{app:texture-probing}, where the bottleneck is matching textures in the executor rather than selecting frames.
Under unseen appearances, the gap becomes larger.
Storing raw images alone is therefore not sufficient. The gain also relies on selecting the right frames.

\FloatBarrier
\section{Coordinate Memory Baseline}
\label{app:coordinate-memory}

Many \benchmark tasks involve positions, which can also be stored as coordinates.
We test when a position alone is enough and when the robot also needs to remember what it saw there.
Following the grounded subgoals in RoboMME~\citep{robomme}, we fine-tune the same Qwen3-VL-4B to output 2D image coordinates as memory instead of selecting keyframes.
Since textures cannot be written as coordinates, for the Appearance tasks the VLM outputs the coordinates of the object to manipulate.
A $\pi_{0.5}$ executor is trained on the same demonstrations and receives the predicted coordinates as text.

\begin{table}[H]
\centering
\caption{\textbf{Coordinate memory baseline.} Success rates on \benchmark under the standard appearance. Both methods use Qwen3-VL-4B to build memory and a $\pi_{0.5}$ executor. Only the memory format differs.}
\label{tab:app-coordinate-memory}
\renewcommand{\arraystretch}{1.3}
\resizebox{\linewidth}{!}{
\begin{tabular}{lcccccccccc}
\toprule
 & Explore & Mine & Hidden & Seal & Spring & Drawer & Flash & Scroll & Placement & Avg. \\
Memory & Cube & Sweeper & Pattern & Checking & Compression & Match & Beacon & Discovery & Imitation & \\
\midrule
Coordinates & 0\% & 46\% & \textbf{34\%} & 13\% & \textbf{68\%} & 43\% & \textbf{48\%} & 0\% & 85\% & 37.4\% \\
Images (\method) & \textbf{2\%} & \textbf{93\%} & 32\% & \textbf{64\%} & 52\% & \textbf{47\%} & 38\% & \textbf{22\%} & \textbf{99\%} & \textbf{49.9\%} \\
\bottomrule
\end{tabular}}
\end{table}

As shown in Table~\ref{tab:app-coordinate-memory}, coordinate memory is lower on average, and the gap depends on the task type.
On the Spatial tasks, only a position is needed. Coordinates work well and outperform \method on two of the three tasks.
On the Spatial \& Appearance tasks, \method performs much better. Each position must be paired with a color. Coordinate memory must get every pairing right when it writes the memory, while image memory keeps both the position and the color.
On the Appearance tasks, the two are close, except that coordinate memory fails on Scroll Discovery. The VLM must match the textures before it can output coordinates, so the difficulty moves from the executor to the VLM.
Coordinates therefore suffice when only a position matters. When positions must be paired with appearances, or the evidence is a texture, image memory performs better.

\FloatBarrier

\end{document}